\documentclass{article}

\usepackage[final, nonatbib]{neurips_2021}

\usepackage[utf8]{inputenc} 
\usepackage[T1]{fontenc}    
\usepackage{hyperref}       
\usepackage{url}            
\usepackage{booktabs}       
\usepackage{amsfonts}       
\usepackage{nicefrac}       
\usepackage{microtype}      
\usepackage{xcolor}         
\usepackage{times}
\usepackage{epsfig}
\usepackage{graphicx}
\usepackage{amsmath}
\usepackage{amssymb}
\usepackage{indentfirst} 
\usepackage{changepage} 
\usepackage{hyperref}
\usepackage{soul}
\usepackage{amsthm}
\usepackage{booktabs}
\usepackage{algorithm}
\usepackage{algorithmic}
\usepackage{makecell}
\usepackage{appendix}
\usepackage{subfigure}
\usepackage{fancyhdr}
\usepackage{authblk}
\usepackage{wrapfig}

\title{Deep Residual Learning in Spiking Neural Networks}

\author{Wei Fang$^{1,2}$, Zhaofei Yu$^{1,2 *}$, Yanqi Chen$^{1,2}$,\\ Tiejun Huang$^{1,2}$, Timothée Masquelier$^{3}$, Yonghong Tian$^{1,2}$\thanks{Corresponding author}\\
~\\
$^{1}$Department of Computer Science and Technology, Peking University\\
$^{2}$Peng Cheng Laboratory, Shenzhen 518055, China\\
$^3$Centre de Recherche Cerveau et Cognition, UMR5549 CNRS - Univ. Toulouse 3 , Toulouse, France\\
}
\begin{document}

\maketitle

\begin{abstract}
	Deep Spiking Neural Networks (SNNs) present optimization difficulties for gradient-based approaches due to discrete binary activation and complex spatial-temporal dynamics. Considering the huge success of ResNet in deep learning, it would be natural to train deep SNNs with residual learning. Previous Spiking ResNet mimics the standard residual block in ANNs and simply replaces ReLU activation layers with spiking neurons, which suffers the degradation problem and can hardly implement residual learning. In this paper, we propose the spike-element-wise (SEW) ResNet to realize residual learning in deep SNNs. We prove that the SEW ResNet can easily implement identity mapping and overcome the vanishing/exploding gradient problems of Spiking ResNet. We evaluate our SEW ResNet on ImageNet, DVS Gesture, and CIFAR10-DVS datasets, and show that SEW ResNet outperforms the state-of-the-art directly trained SNNs in both accuracy and time-steps.  Moreover, SEW ResNet can achieve higher performance by simply adding more layers, providing a simple method to train deep SNNs. To our best knowledge, this is the first time that directly training deep SNNs with more than 100 layers becomes possible. Our codes are available at \url{https://github.com/fangwei123456/Spike-Element-Wise-ResNet}.

\end{abstract}

\section{Introduction}
Artificial Neural Networks (ANNs) have achieved great success in many tasks, including image classification~\cite{krizhevsky2012imagenet, simonyan2015deep, szegedy2015going}, object detection~\cite{girshick2014rich, liu2016ssd, redmon2016you}, 
machine translation~\cite{bahdanau2014neural}, and gaming~\cite{mnih2015human, silver2016mastering}. One of the critical factors for ANNs' success is deep learning~\cite{lecun2015deep}, which uses multi-layers to learn representations of data with multiple levels of abstraction. It has been proved that deeper networks have advantages over shallower networks in computation cost and generalization ability~\cite{bengio2007scaling}. The function represented by a deep network can require an exponential number of hidden units by a shallow network with one hidden layer~\cite{NIPS2014_109d2dd3}. 
In addition, the depth of the network is closely related to the network's performance in practical tasks~\cite{simonyan2015deep, szegedy2015going,krizhevsky2014weird, simonyan2015deep}. 
Nevertheless, recent evidence~\cite{he2015convolutional, srivastava2015highway, he2015deep} reveals that with the network depth increasing, the accuracy gets saturated and then degrades rapidly. To solve this degradation problem, residual learning is proposed \cite{he2015deep,he2016identity} and the residual structure is widely exploited in ``very deep'' networks that achieve the leading performance~\cite{i2016squeezenet, xie2016aggregated, huang2017densely, vaswani2017attention}.


Spiking Neural Networks (SNNs) are regarded as a potential competitor of ANNs for their high biological plausibility, event-driven property, and low power consumption~\cite{roy2019towards}. Recently, deep learning methods are introduced into SNNs, and deep SNNs have achieved close performance as ANNs in some simple classification datasets~\cite{TAVANAEI201947}, but still worse than ANNs in complex tasks, e.g., classifying the ImageNet dataset~\cite{russakovsky2015imagenet}. To obtain higher performance SNNs, it would be natural to explore deeper network structures like ResNet.  Spiking ResNet~\cite{KIM2018373, 10.1007/978-3-030-36718-3_15, 10.3389/fnins.2021.629000, hu2020spiking, sengupta2019going, Han_2020_CVPR,lee2020enabling, zheng2020going, samadzadeh2021convolutional,rathi2020dietsnn, rathi2020enabling}, as the spiking version of ResNet, is proposed by mimicking the residual block in ANNs and replacing ReLU activation layers with spiking neurons. Spiking ResNet converted from ANN achieves state-of-the-art accuracy on nearly all datasets, while the directly trained Spiking ResNet has not been validated to solve the degradation problem.


In this paper, we show that Spiking ResNet is inapplicable to all neuron models to achieve identity mapping. Even if the identity mapping condition is met, Spiking ResNet suffers from the problems of vanishing/exploding gradient. Thus, we propose the Spike-Element-Wise (SEW) ResNet to realize residual learning in SNNs. We prove that the SEW ResNet can easily implement identity mapping and overcome the vanishing/exploding gradient problems at the same time. 
We evaluate Spiking ResNet and SEW ResNet on both the static ImageNet dataset and the neuromorphic DVS Gesture dataset~\cite{amir2017low}, CIFAR10-DVS dataset~\cite{10.3389/fnins.2017.00309}. The experiment results are consistent with our analysis, indicating that the deeper Spiking ResNet suffers from the degradation problem --- the deeper network has higher training loss than the shallower network, while SEW ResNet can achieve higher performance by simply increasing the network’s depth.
Moreover, we show that SEW ResNet outperforms the state-of-the-art directly trained SNNs in both accuracy and time-steps. To the best of our knowledge, this is the first time to explore the directly-trained deep SNNs with more than 100 layers.



\section{Related Work}
\subsection{Learning Methods of Spiking Neural Networks}
ANN to SNN conversion (ANN2SNN)~\cite{hunsberger2015spiking, cao2015spiking, Bodo2017Conversion, sengupta2019going, Han_2020_CVPR, han2020deep, deng2021optimal, stockl2021optimized, pmlr-v139-li21d} and backpropagation with surrogate gradient ~\cite{neftci2019surrogate} are the two main methods to get deep SNNs. The ANN2SNN method firstly trains an ANN with ReLU activation, then converts the ANN to an SNN by replacing ReLU with spiking neurons and adding scaling operations like weight normalization and threshold balancing. Some recent conversion methods have achieved near loss-less accuracy with VGG-16 and ResNet~\cite{Han_2020_CVPR, han2020deep, deng2021optimal, pmlr-v139-li21d}. However, the converted SNN needs a longer time to rival the original ANN in precision as the conversion is based on rate-coding~\cite{Bodo2017Conversion}, which increases the SNN's latency and restricts the practical application. The backpropagation methods can be classified into two categories~\cite{kim2020unifying}. The method in the first category computes the gradient by unfolding the network over the simulation time-steps~\cite{lee2016training, huh2017gradient, wu2018STBP, shrestha2018slayer, lee2020enabling, neftci2019surrogate}, which is similar to the idea of backpropagation through time (BPTT). As the gradient with respect to the threshold-triggered firing is non-differentiable, the surrogate gradient is often used. The SNN trained by the surrogate method is not limited to rate-coding, and can also be applied on temporal tasks, e.g., classifying neuromorphic datasets~\cite{wu2018STBP, fang2020incorporating, HE2020108}. The second method computes the gradients of the timings of existing spikes with respect to the membrane potential at the spike timing \cite{comsa2020temporal, mostafa2017supervised, kheradpisheh2020temporal, zhou2019temporal, zhang2020temporal}.


\subsection{Spiking Residual Structure}
Previous ANN2SNN methods noticed the distinction between plain feedforward ANNs and residual ANNs, and made specific normalization for conversion. Hu et al.~\cite{hu2020spiking} were the first to apply the residual structure in ANN2SNN with scaled shortcuts in SNN to match the activations of the original ANN. Sengupta et al.~\cite{sengupta2019going} proposed Spike-Norm to balance SNN's threshold and verified their method by converting VGG and ResNet to SNNs. Existing backpropagation-based methods use nearly the same structure from ResNet. Lee et al.~\cite{lee2020enabling} evaluated their custom surrogate methods on shallow ResNets whose depths are no more than ResNet-11. Zheng et al.~\cite{zheng2020going} proposed the threshold-dependent batch normalization (td-BN) to replace naive batch normalization (BN) \cite{BN} and successfully trained Spiking ResNet-34 and Spiking ResNet-50 directly with surrogate gradient by adding td-BN in shortcuts.

\section{Methods}
\subsection{Spiking Neuron Model}
The spiking neuron is the fundamental computing unit of SNNs. Similar to Fang et al.~\cite{fang2020incorporating},  
we use a unified model to describe the dynamics of all
kinds of spiking neurons, which includes the following discrete-time equations:
\begin{align}
	H[t] &= f(V[t - 1], X[t]), \label{neural dynamics}\\
	S[t] &= \Theta(H[t] - V_{th}), \label{neural spiking}\\
	V[t] &= H[t]~\left( 1 - S[t] \right) + V_{reset}~S[t], \label{neural reset}
\end{align}
where $X[t]$ is the input current at time-step $t$, $H[t]$ and $V[t]$ denote the membrane potential after neuronal dynamics and after the trigger of a spike at time-step $t$, respectively.
$V_{th}$ is the firing threshold,  $\Theta(x)$ is the Heaviside step function and is defined by $\Theta(x) = 1$ for $x \ge 0$ and $\Theta(x) = 0$ for $x < 0$. $S[t]$ is the output spike at time-step $t$,  which equals 1 if there is a spike and 0 otherwise. $V_{reset}$ denotes the reset potential. The function $f(\cdot)$ in Eq.~\eqref{neural dynamics} describes the neuronal dynamics and takes different forms for different spiking neuron models. For example, the function $f(\cdot)$ for the Integrate-and-Fire (IF) model and Leaky Integrate-and-Fire (LIF) model can be described by Eq.~\eqref{IFneuron} and Eq.~\eqref{LIFneuron}, respectively.
\begin{align}
	H[t]&=V[t - 1]+X[t], \label{IFneuron}\\
	H[t]&=V[t - 1]+\frac{1}{\tau}(X[t]-(V[t - 1]-V_{reset})),   \label{LIFneuron} 
\end{align}
where $\tau$ represents the membrane time constant. Eq.~\eqref{neural spiking} and Eq.~\eqref{neural reset} describe the spike generation and resetting processes, which are the same for all kinds of spiking neuron models. In this paper, the surrogate gradient method is used to define $\Theta'(x) \triangleq \sigma'(x)$ during error back-propagation, with $\sigma(x)$ denoting the surrogate function.   

\subsection{Drawbacks of Spiking ResNet}
\begin{figure}
	\begin{center}
		\includegraphics[width=0.7\textwidth,trim=0 136 280 0,clip]{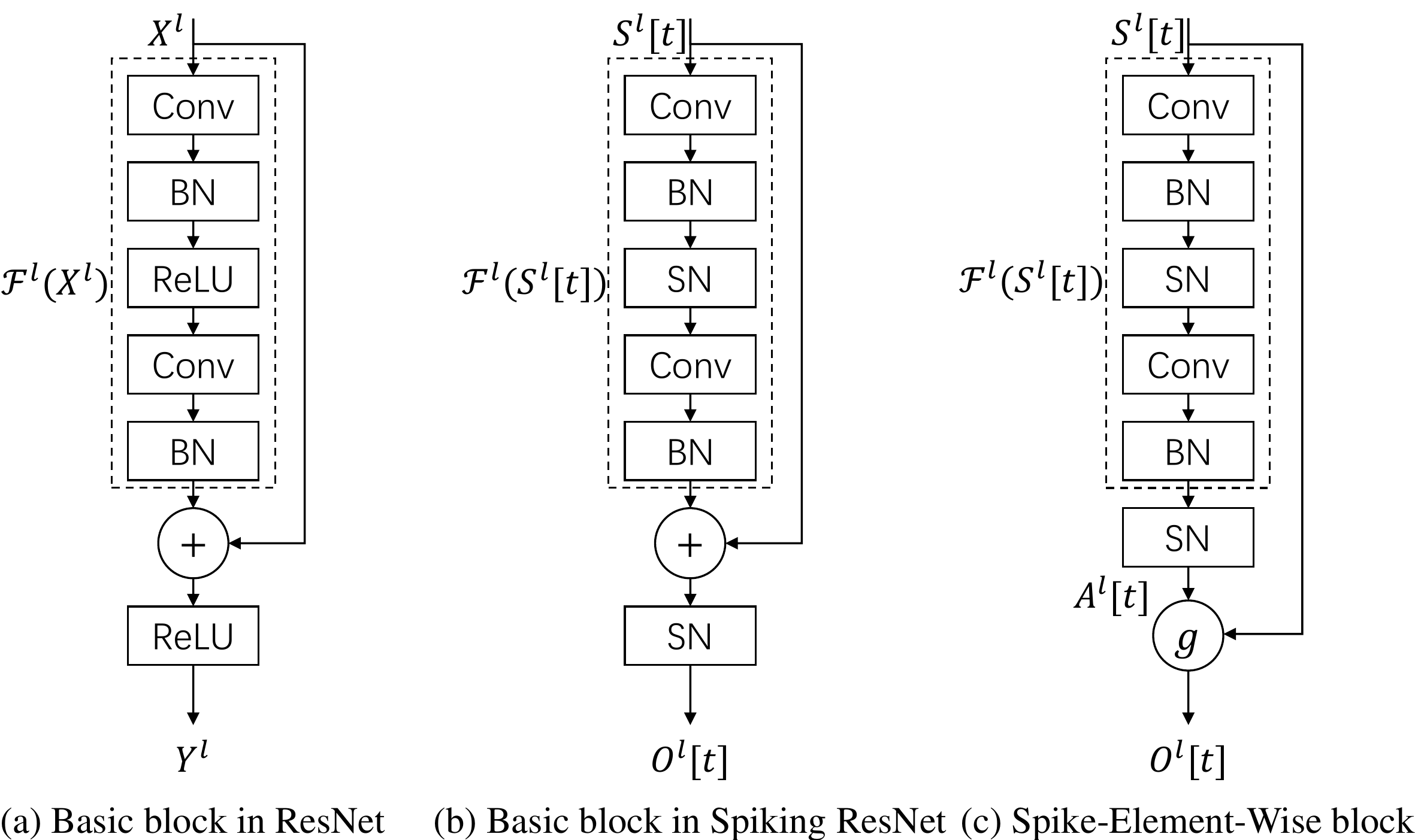}
		\vspace{-0.1cm}
		\caption{Residual blocks in ResNet, Spiking ResNet and SEW ResNet.} 
		\vspace{-0.5cm}
		\label{figure:basic block}
	\end{center}
\end{figure}
The residual block is the key component of ResNet. Fig.~\ref{figure:basic block}(a) shows the basic block in ResNet \cite{he2015deep}, where $X^{l}, Y^{l}$ are the input and output of the $l$-th block in  ResNet, Conv is the convolutional layer, BN denotes batch normalization, and ReLU denotes the rectified linear unit activation layer. The basic block of Spiking ResNet used in \cite{zheng2020going, hu2020spiking, lee2020enabling} simply mimics the block in ANNs by replacing ReLU activation layers with spiking neurons (SN), which is illustrated in Fig.~\ref{figure:basic block}(b). Here $S^{l}[t], O^{l}[t]$ are the input and output of the $l$-th block in Spiking ResNet at time-step $t$. Based on the above definition, 
we will analyze the drawbacks of Spiking ResNet below.

\textbf{Spiking ResNet is inapplicable to all neuron models to achieve identity mapping.} 
One of the critical concepts in ResNet is identity mapping. He et al.~\cite{he2015deep} noted that if the added layers implement the identity mapping, a deeper model should have training error no greater than its shallower counterpart. However, it is unable to train the added layers to implement identity mapping in a feasible time, resulting in deeper models performing worse than shallower models (the degradation problem). To solve this problem, the residual learning is proposed by adding a shortcut connection (shown in Fig.~\ref{figure:basic block}(a)). If we use $\mathcal{F}^{l}$ to denote the residual mapping, e.g., a stack of two convolutional layers, of the $l$-th residual block in ResNet and Spiking ResNet, then the residual block in Fig.\ref{figure:basic block}(a) and Fig.\ref{figure:basic block}(b) can be formulated as 
\begin{align}
	Y^{l} &= {\rm ReLU}(\mathcal{F}^{l}(X^{l}) + X^{l}), \label{block ann}\\
	O^{l}[t] &= {\rm SN}(\mathcal{F}^{l}(S^{l}[t]) + S^{l}[t]). \label{block snn}
\end{align}
The  residual block of Eq.~\eqref{block ann} make it easy to implement identity mapping in ANNs. To see this, when $\mathcal{F}^{l}(X^{l}) \equiv 0$,  $Y^{l} = \rm ReLU(X^{l})$. In most cases, $X^{l}$ is the activation of the previous ReLU layer and $X^{l} \geq 0$. Thus, $Y^{l} = {\rm ReLU}(X^{l}) = X^{l}$, which is identity mapping.

Different from ResNet, the residual block in Spiking ResNet (Eq.~(\ref{block snn}))
restricts the models of spiking neuron to implement identity mapping. When $\mathcal{F}^{l}(S^{l}[t]) \equiv 0$, $O^{l}[t] = {\rm SN}(S^{l}[t]) \neq S^{l}[t]$. To transmit $S^{l}[t]$ and make ${\rm SN}(S^{l}[t]) = S^{l}[t]$, the last spiking neuron (SN) in the $l$-th residual block needs to fire a spike after receiving a spike, and keep silent after receiving no spike at time-step $t$.
It works for IF neuron described by Eq.~\eqref{IFneuron}. Specifically, we can set $0 < V_{th} \leq 1$ and $V[t-1]=0$ to ensure that $X[t]=1$ leads to $H[t] \geq V_{th}$, and $X[t]=0$ leads to $H[t] < V_{th}$.
However, when considering some spiking  neuron models with complex neuronal dynamics, it is hard to achieve ${\rm SN}(S^{l}[t])=S^{l}[t]$. For example, the LIF neuron used in \cite{zimmer2019technical, fang2020incorporating, yin2020effective} considers a learnable membrane time constant $\tau$, the neuronal dynamics of which can be described with Eq.~\eqref{LIFneuron}.
When $X[t] = 1$ and $V[t-1]=0$, $H[t] = \frac{1}{\tau}$. It is difficult to find a firing threshold that ensures  $H[t]  > V_{th}$ as $\tau$ is being changed in training by the optimizer.


\textbf{Spiking ResNet suffers from the problems of vanishing/exploding gradient.} \label{Spiking ResNet suffers from the problems of vanishing/exploding gradient}
Consider a spiking ResNet with $k$ sequential blocks to transmit $S^{l}[t]$, and the identity mapping condition is met, e.g., the spiking neurons are the IF neurons with $0 < V_{th} \leq 1$, then we have $S^{l}[t] = S^{l+1}[t] = ... = S^{l+k-1}[t] = O^{l+k-1}[t]$. Denote the $j$-th element in $S^{l}[t]$ and $O^{l}[t]$ as $S_{j}^{l}[t]$ and $O_{j}^{l}[t]$ respectively, the gradient of the output of the $(l+k-1)$-th residual block with respect to the input of the $l$-th residual block can be calculated layer by layer:
\begin{align}
	\frac{\partial O_{j}^{l+k-1}[t]}{\partial S_{j}^{l}[t]} = \prod_{i=0}^{k-1} \frac{\partial O_{j}^{l+i}[t]}{\partial S_{j}^{l+i}[t]}  = \prod_{i=0}^{k-1}\Theta'(S_{j}^{l+i}[t] - V_{th})  \to 
	\begin{cases}
		0, \text{if}~0 < \Theta'(S_{j}^{l}[t] - V_{th}) < 1 \\
		1, \text{if}~\Theta'(S_{j}^{l}[t] - V_{th}) = 1 \\
		+\infty, \text{if}~\Theta'(S_{j}^{l}[t] - V_{th}) > 1
	\end{cases},
	\label{block snn gradient}
\end{align}
where $\Theta(x)$ is the Heaviside step function and $\Theta'(x)$ is defined by the surrogate gradient. The second equality hold as $O_{j}^{l+i}[t]={\rm SN}(S_{j}^{l+i}[t])$. In view of the fact that $S_{j}^{l}[t]$ can only take $0$ or $1$,  $\Theta'(S_{j}^{l}[t] - V_{th}) =1$ is not satisfied for commonly used surrogate functions mentioned in \cite{neftci2019surrogate}. Thus, the vanishing/exploding gradient problems are prone to happen in deeper Spiking ResNet.

Based on the above analysis, we believe that the previous Spiking ResNet ignores the highly nonlinear caused by spiking neurons, and can hardly implement residual learning. Nonetheless,
the basic block in Fig.~\ref{figure:basic block}(b) is still decent for ANN2SNN with extra normalization \cite{hu2020spiking, sengupta2019going}, as the SNN converted from ANN aims to use firing rates to match the origin ANN's activations.

\subsection{Spike-Element-Wise ResNet} \label{Spike-Element-Wise ResNet}
Here we propose the Spike-Element-Wise (SEW) residual block to realize the residual learning in SNNs, which can easily implement identity mapping and overcome the vanishing/exploding gradient problems at the same time.  As illustrated in Fig.~\ref{figure:basic block}(c), the SEW residual block can be formulated as: 
\begin{align}
	O^{l}[t] = g({\rm SN}(\mathcal{F}^{l}(S^{l}[t])), S^{l}[t])=g(A^{l}[t], S^{l}[t])\label{spike-element-wise residual block},
\end{align}
where $g$ represents an element-wise function with two spikes tensor as inputs. Here we use $A^{l}[t]$ to denote the residual mapping to be learned as $A^{l}[t]= {\rm SN}(\mathcal{F}^{l}(S^{l}[t]))$.

\begin{wraptable}{r}{6cm}
	\centering
	\scalebox{0.8}
	{
		\begin{tabular}{cc}
			\hline
			\textbf{Name} &\textbf{Expression of $g(A^{l}[t], S^{l}[t])$}\\
			\hline
			ADD & $A^{l}[t] + S^{l}[t]$\\
			AND & $A^{l}[t] \land S^{l}[t]$ = $A^{l}[t] \cdot S^{l}[t]$\\
			IAND & $(\neg A^{l}[t]) \land S^{l}[t]$ = $ (1 - A^{l}[t]) \cdot S^{l}[t]$\\
			\hline
		\end{tabular}
	}
	\caption{List of element-wise functions $g$.} 
	\label{tab:g}
	\vspace{-0.4cm}
\end{wraptable}

\textbf{SEW ResNet can easily implement identity mapping.} 
By utilizing the binary property of spikes, we can find different element-wise functions $g$ that satisfy identity mapping (shown in Tab.~\ref{tab:g}). To be specific, when choosing \textit{ADD} and \textit{IAND} as element-wise functions $g$, identity mapping is achieved by setting $A^{l}[t] \equiv 0$, which can be implemented simply by setting the weights and the bias of the last batch normalization layer (BN) in $\mathcal{F}^{l}$ to zero. Then we can get $O^{l}[t] = g(A^{l}[t], S^{l}[t])=g({\rm SN}(0), S^{l}[t]) = g(0, S^{l}[t]) = S^{l}[t]$. This is applicable to all neuron models. When using \textit{AND} as the element-wise function $g$, we set $A^{l}[t] \equiv 1$ to get identity mapping. It can be implemented by setting the last BN's weights to zero and the bias to a large enough constant to cause spikes, e.g., setting the bias as $V_{th}$ when the last $\rm SN$ is IF neurons. Then we have $O^{l}[t] = 1 \land S^{l}[t] = S^{l}[t]$. Note that using \textit{AND} may suffer from the same problem as Spiking ResNet. It is hard to control some spiking neuron models with complex neuronal dynamics to generate spikes at a specified time-step. 

\begin{figure}
	\begin{center}
		\subfigure[Downsample basic block]{\includegraphics[width=0.3\textwidth,trim=100 310 600 0,clip]{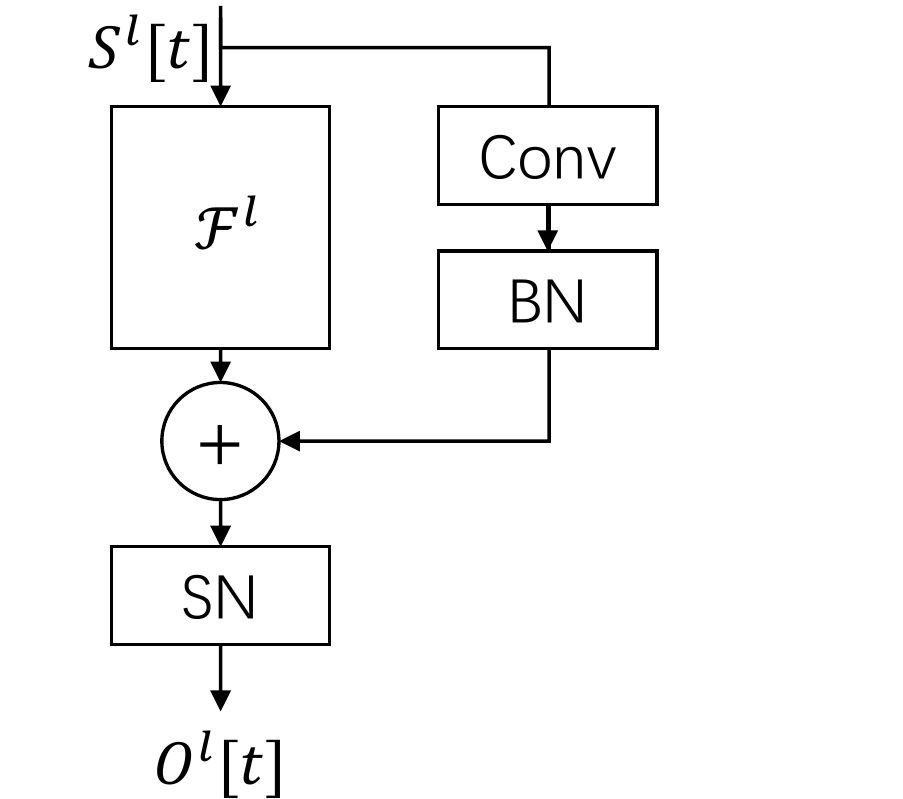}}\hspace{2cm}
		\subfigure[Downsample SEW block]{\includegraphics[width=0.3\textwidth,trim=500 310 200 0,clip]{./fig/block2.pdf}}
		\vspace{-0.1cm}
		\caption{Downsample blocks in Spiking ResNet and SEW ResNet.}
		\vspace{-0.5cm}
		\label{figure:downsample block}
	\end{center}
\end{figure}

\textbf{Formulation of downsample block.} Remarkably, when the input and output of one block have different dimensions, the shortcut is set as convolutional layers with stride $>1$, rather than the identity connection, to perform downsampling. The ResNet and the Spiking ResNet utilize \{Conv-BN\} without ReLU in shortcut (Fig.~\ref{figure:downsample block}(a)).
In contrast, we add a SN in shortcut (Fig.~\ref{figure:downsample block}(b)).

\textbf{SEW ResNet can overcome vanishing/exploding gradient.}
The SEW block is similar to \textit{ReLU before addition} (RBA) block \cite{he2016identity} in ANNs, which can be formulated as
\begin{align}
	Y^{l} &= {\rm ReLU}(\mathcal{F}^{l}(X^{l})) +  X^{l}.\label{ReLU before addition block}
\end{align}
The RBA block is criticized by He et al.~\cite{he2016identity} for $X^{l+1} = Y^{l} \geq X^{l}$, which will cause infinite outputs in deep layers. The experiment results in \cite{he2016identity} also showed that the performance of the RBA block is worse than the basic block (Fig.\ref{figure:basic block}(a)). To some extent, the SEW block is an extension of the RBA block. Note that using \textit{AND} and \textit{IAND} as $g$ will output spikes (i.e. binary tensors), which means that the infinite outputs problem in ANNs will never occur in SNNs with SEW blocks, since all spikes are less or equal than 1. 
When choosing \textit{ADD} as $g$, the infinite outputs problem can be relieved as the output of $k$ sequential SEW blocks will be no larger than $k+1$. In addition, a downsample SEW block will regulate the output to be no larger than 2 when $g$ is \textit{ADD}.

When the identity mapping is implemented, the gradient of the output of the $(l+k-1)$-th SEW block with respect to the input of the $l$-th SEW block can be calculated layer by layer: 
\begin{align}
	\frac{\partial O_{j}^{l+k-1}[t]}{\partial S_{j}^{l}[t]}  = \prod_{i=0}^{k-1}\frac{\partial g(A_{j}^{l+i}[t], S_{j}^{l+i}[t])}{\partial S_{j}^{l+i}[t]} = 
	\begin{cases}
		\prod_{i=0}^{k-1}\frac{\partial (0 + S_{j}^{l+i}[t])}{\partial S_{j}^{l+i}[t]}, \text{if}~~g = ADD \\
		\prod_{i=0}^{k-1}\frac{\partial (1 \cdot S_{j}^{l+i}[t])}{\partial S_{j}^{l+i}[t]}, \text{if}~~g = AND \\
		\prod_{i=0}^{k-1}\frac{\partial ((1 - 0) \cdot S_{j}^{l+i}[t])}{\partial S_{j}^{l+i}[t]}, \text{if}~~g = IAND
	\end{cases}
	= 1.
	\label{sew block snn gradient}
\end{align}
The second equality holds as identity mapping is achieved by setting $A^{l+i}[t] \equiv 1$ for $g=$ \textit{AND}, and $A^{l+i}[t] \equiv 0$ for $g=$ \textit{ADD/IAND}.
Since the gradient in Eq.~\eqref{sew block snn gradient} is a constant, the SEW ResNet can overcome the vanishing/exploding gradient problems.


\section{Experiments}
\subsection{ImageNet Classification}
As the test server of ImageNet 2012 is no longer available,  we can not report the actual test accuracy. Instead, we use the accuracy on the \textit{validation} set as the test accuracy, which is the same as \cite{hu2020spiking, zheng2020going}.
He et al.~\cite{he2015deep} evaluated the 18/34/50/101/152-layer ResNets on the ImageNet dataset. For comparison, we consider the SNNs with the same network architectures, except that the basic residual block (Fig.\ref{figure:basic block}(a)) is replaced by the spiking basic block (Fig.\ref{figure:basic block}(b)) and SEW block (Fig.\ref{figure:basic block}(c)) with $g$ as \textit{ADD}, respectively. We denote the SNN with the basic block as \textit{Spiking ResNet} and the SNN with the SEW block as \textit{SEW ResNet}. The IF neuron model is adopted for the static ImageNet dataset. During training on ImageNet, we find that the Spiking ResNet-50/101/152 can not converge unless we use the zero initialization~\cite{goyal2018accurate}, which sets all blocks to be an identity mapping at the start of training. Thus, the results of Spiking ResNet-18/34/50/101/152 reported in this paper are with zero initialization.


\textbf{Spiking ResNet \emph{vs.} SEW ResNet.}
\begin{figure}
	\begin{center}
		\subfigure[Training loss]{\includegraphics[width=0.49\textwidth,trim=0 255 510 0,clip]{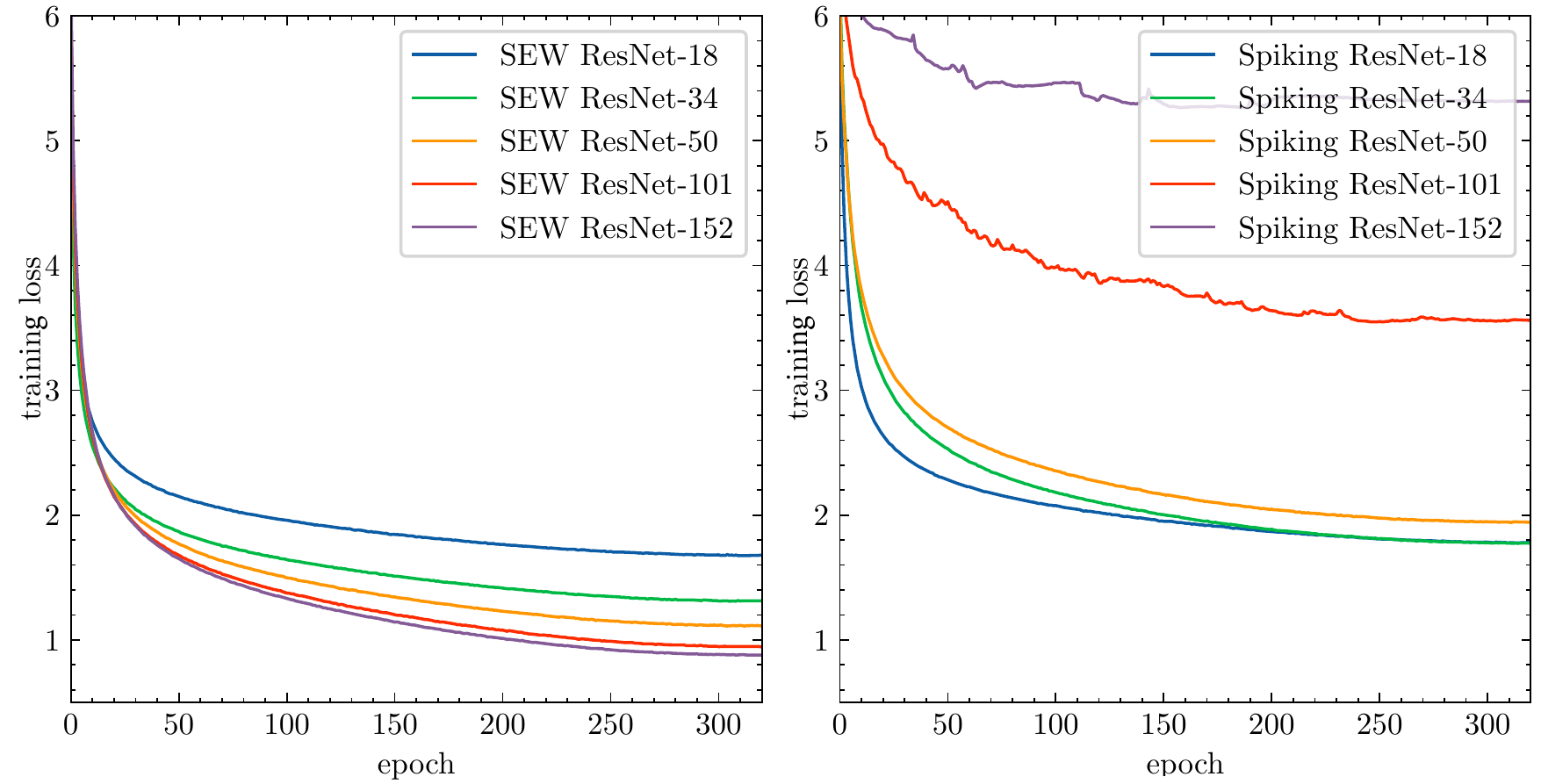}}
		\subfigure[Training accuracy]{\includegraphics[width=0.49\textwidth,trim=510 255 0 0,clip]{./fig/fig3_1.pdf}}
		\subfigure[Test accuracy]{\includegraphics[width=0.49\textwidth,trim=255 0 255 255,clip]{./fig/fig3_1.pdf}}
		\vspace{-0.25cm}
		\caption{Comparison of the training loss, training accuracy and test accuracy on ImageNet.} 
		\label{figure:cmp on ImageNet}
	\end{center}
\end{figure}
We first evaluate the performance of Spiking ResNet and SEW ResNet. Tab.~\ref{tab:acc on ImageNet} reports the test accuracy on ImageNet validation. The results show that the deeper 34-layer Spiking ResNet has lower test accuracy than the shallower 18-layer Spiking ResNet. 
\begin{table}
	\centering
	\scalebox{0.9}
	{
		\begin{tabular}{ccccc}
			\hline
			\textbf{Network} & \multicolumn{2}{c}{\textbf{SEW ResNet (ADD)}} & \multicolumn{2}{c}{\textbf{Spiking ResNet}}\\
			& \textbf{Acc@1(\%)} & \textbf{Acc@5(\%)} & \textbf{Acc@1(\%)} & \textbf{Acc@5(\%)}\\
			\hline
			ResNet-18 &63.18 &84.53 &62.32 &84.05 \\
			ResNet-34 &67.04 &87.25 &61.86 &83.69 \\
			ResNet-50 &67.78 &87.52 &57.66 &80.43 \\
			ResNet-101 &68.76 &88.25 &31.79 &54.91 \\
			ResNet-152 &69.26 &88.57 &10.03 &23.57\\
			\hline
		\end{tabular}
	}
	\caption{Test accuracy on ImageNet.}
	\label{tab:acc on ImageNet}
	\vspace{-0.8cm}
\end{table}
As the layer increases, the test accuracy of Spiking ResNet decreases. To reveal the reason, we compare the training loss, training accuracy, and test accuracy of Spiking ResNet during the training procedure, which is shown in Fig.~\ref{figure:cmp on ImageNet}. We can find the degradation problem of the Spiking ResNet --- the deeper network has higher \textit{training loss} than the shallower network. In contrast, the deeper 34-layer SEW ResNet has higher test accuracy than the shallower 18-layer SEW ResNet (shown in Tab.~\ref{tab:acc on ImageNet}). More importantly, it can be found from Fig.~\ref{figure:cmp on ImageNet} that the training loss of our SEW ResNet decreases and the training/test accuracy increases with the increase of depth, which indicates that we can obtain higher performance by simply increasing the network's depth. All these results imply that the degradation problem is well addressed by SEW ResNet.


\textbf{Comparisons with State-of-the-art Methods.}
In Tab.~\ref{tab:cmp sota on ImageNet}, we compare SEW ResNet with previous Spiking ResNets that achieve the best results on ImageNet. To our best knowledge, the SEW ResNet-101 and the SEW ResNet-152 are the only SNNs with more than 100 layers to date, and there are no other networks with the same structure to compare. When the network structure is the same, our SEW ResNet outperforms the state-of-the-art accuracy of directly trained Spiking ResNet, even with fewer time-steps $T$. The accuracy of SEW ResNet-34 is slightly lower than Spiking ResNet-34 (large) with td-BN (67.04\% v.s. 67.05\%), which uses 1.5 times as many simulating time-steps $T$ (6 v.s. 4) and 4 times as many the number of parameters (85.5M v.s. 21.8M), compared with our SEW ResNet. The state-of-the-art ANN2SNN methods \cite{pmlr-v139-li21d, hu2020spiking} have better accuracy than our SEW ResNet, but they respectively use 64 and 87.5 times as many time-steps as ours.
\begin{table}
	\centering
	\scalebox{0.8}
	{
		\begin{tabular}{llll}
			\hline
			\textbf{Network} & \textbf{Methods} &\textbf{Accuracy(\%)} &  $\boldsymbol {T}$\\
			\hline
			\textbf{SEW ResNet-34} &\textbf{Spike-based BP} &\textbf{67.04} & \textbf{4}\\
			Spiking ResNet-34(large)$^{\dagger}$ with td-BN \cite{zheng2020going} &Spike-based BP & 67.05 & 6\\
			Spiking ResNet-34 with td-BN \cite{zheng2020going} &Spike-based BP & 63.72  & 6\\
			Spiking ResNet-34 \cite{Han_2020_CVPR} &ANN2SNN & 69.89 &4096\\
			Spiking ResNet-34 \cite{sengupta2019going} &ANN2SNN & 65.47 &2000 \\
			Spiking ResNet-34 \cite{pmlr-v139-li21d} & ANN2SNN & 74.61 &256 \\
			Spiking ResNet-34 \cite{rathi2020enabling} &ANN2SNN and Spike-based BP & 61.48 & 250\\
			\hline
			\textbf{SEW ResNet-50} &\textbf{Spike-based BP} &\textbf{67.78} & \textbf{4}\\
			Spiking ResNet-50 with td-BN \cite{zheng2020going} &Spike-based BP & 64.88  & 6\\
			Spiking ResNet-50 \cite{hu2020spiking} &ANN2SNN & 72.75 &350\\
			\hline
			\textbf{SEW ResNet-101} &\textbf{Spike-based BP} &\textbf{68.76} & \textbf{4}\\
			\hline
			\textbf{SEW ResNet-152} &\textbf{Spike-based BP} &\textbf{69.26} & \textbf{4}\\
			
			\hline
		\end{tabular}
		
	}
	\caption{Comparison with previous Spiking ResNet on ImageNet. $\dagger$ has the same network structure as the standard Spiking ResNet-34, but uses four times as many the number of convolution kernels.}
	\vspace{-0.6cm}
	\label{tab:cmp sota on ImageNet}
\end{table}

\begin{figure}
	\begin{center}
		\subfigure{\includegraphics[width=0.49\textwidth,trim=0 0 0 0,clip]{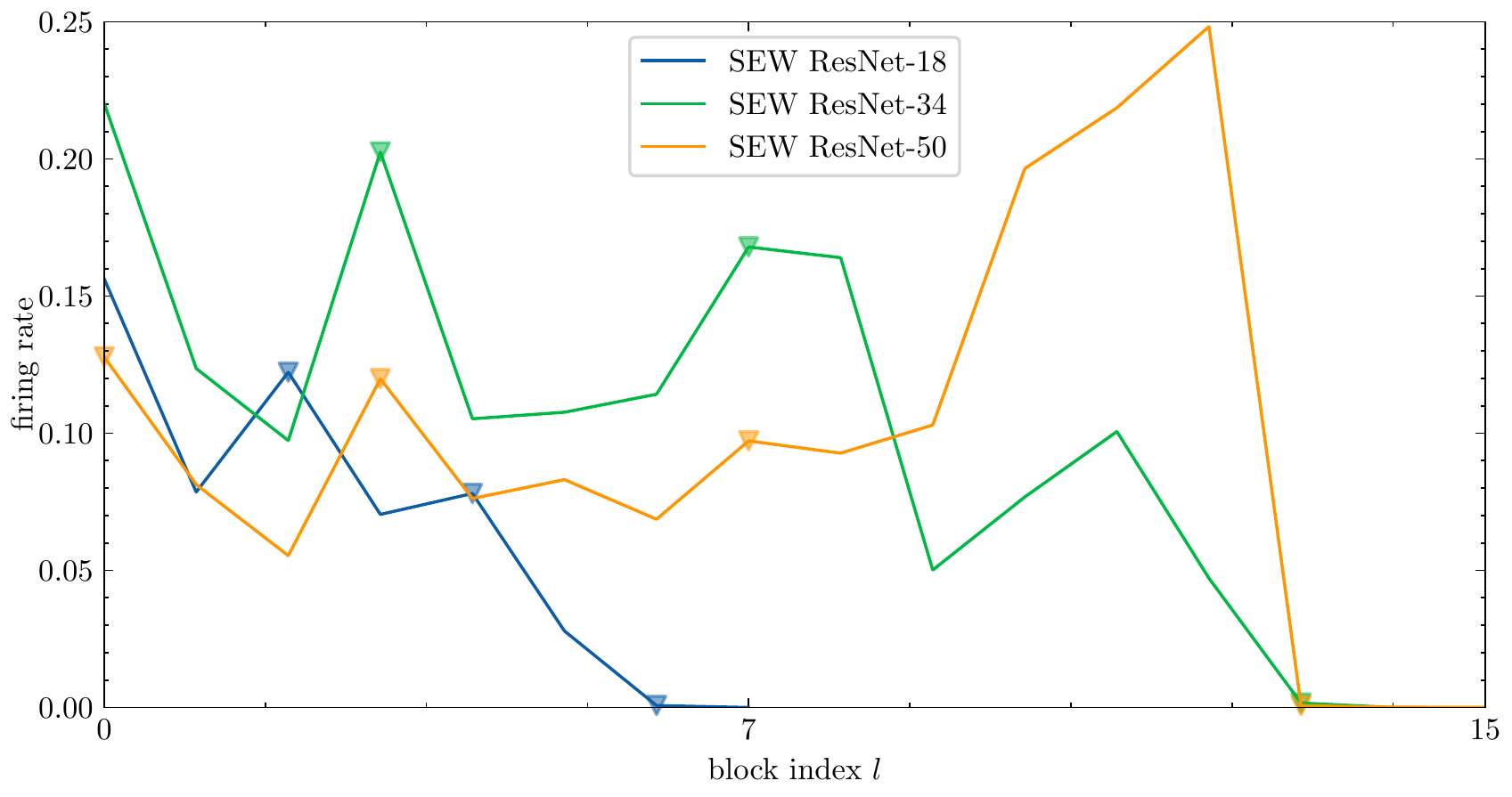}}
		\subfigure{\includegraphics[width=0.49\textwidth,trim=0 2 0 0,clip]{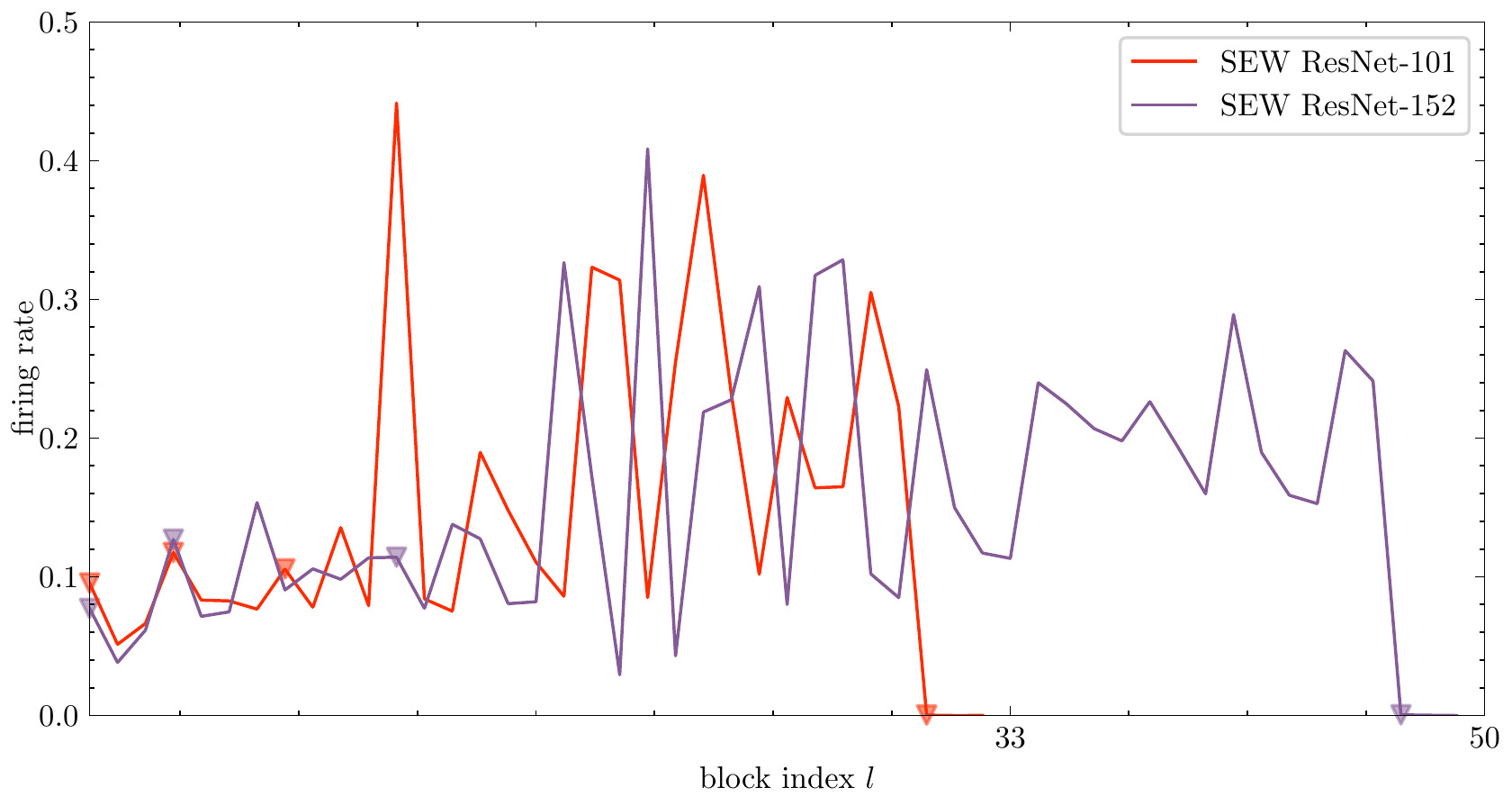}}
		\caption{Firing rates of $A^{l}$ in SEW blocks on ImageNet.} 
		\label{figure:sew block fr on ImageNet}
		\vspace{-0.8cm}
	\end{center}
\end{figure}

\textbf{Analysis of spiking response of SEW blocks.}
Fig.~\ref{figure:sew block fr on ImageNet} shows the firing rates of $A^{l}$ in SEW ResNet-18/34/50/101/152 on ImageNet. There are 7 blocks in SEW ResNet-18, 15 blocks in SEW ResNet-34 and SEW ResNet-50, 33 blocks in SEW ResNet-101, and 50 blocks in SEW ResNet-152. The downsample SEW blocks are marked by the triangle down symbol $\triangledown$. 
As we choose \textit{ADD} as element-wise functions $g$, a lower firing rate means that the SEW block gets closer to implementing identity mapping, except for downsample blocks. Note that the shortcuts of downsample blocks are not identity mapping, which is illustrated in Fig.~\ref{figure:downsample block}(b). Fig.~\ref{figure:sew block fr on ImageNet} shows that all spiking neurons in SEW blocks have low firing rates, and the spiking neurons in the last two blocks even have firing rates of almost zero. As the time-steps $T$ is 4 and firing rates are no larger than 0.25, all neurons in SEW ResNet-18/34/50 fire on average no more than one spike during the whole simulation. Besides, all firing rates in SEW ResNet-101/152 are not larger than 0.5, indicating that all neurons fire on average not more than two spikes. In general, the firing rates of $A^{l}$ in SEW blocks are at a low level, verifying that most SEW blocks act as identity mapping.


\textbf{Gradients Check on ResNet-152 Structure.} \label{Gradients Check on ResNet-152 Structure.}
Eq.~(\ref{block snn gradient}) and Eq.~(\ref{sew block snn gradient}) analyze the gradients of multiple blocks with identity mapping. To verify that SEW ResNet can overcome vanishing/exploding gradient, we check the gradients of Spiking ResNet-152 and SEW ResNet-152, which are the deepest standard ResNet structure. We consider the same initialization parameters and with/without zero initialization.

\begin{figure}[h]
	\begin{center}
		\subfigure[Firing rate of output $O^{l}$]{\includegraphics[width=0.3\textwidth,trim=0 0 560 0,clip]{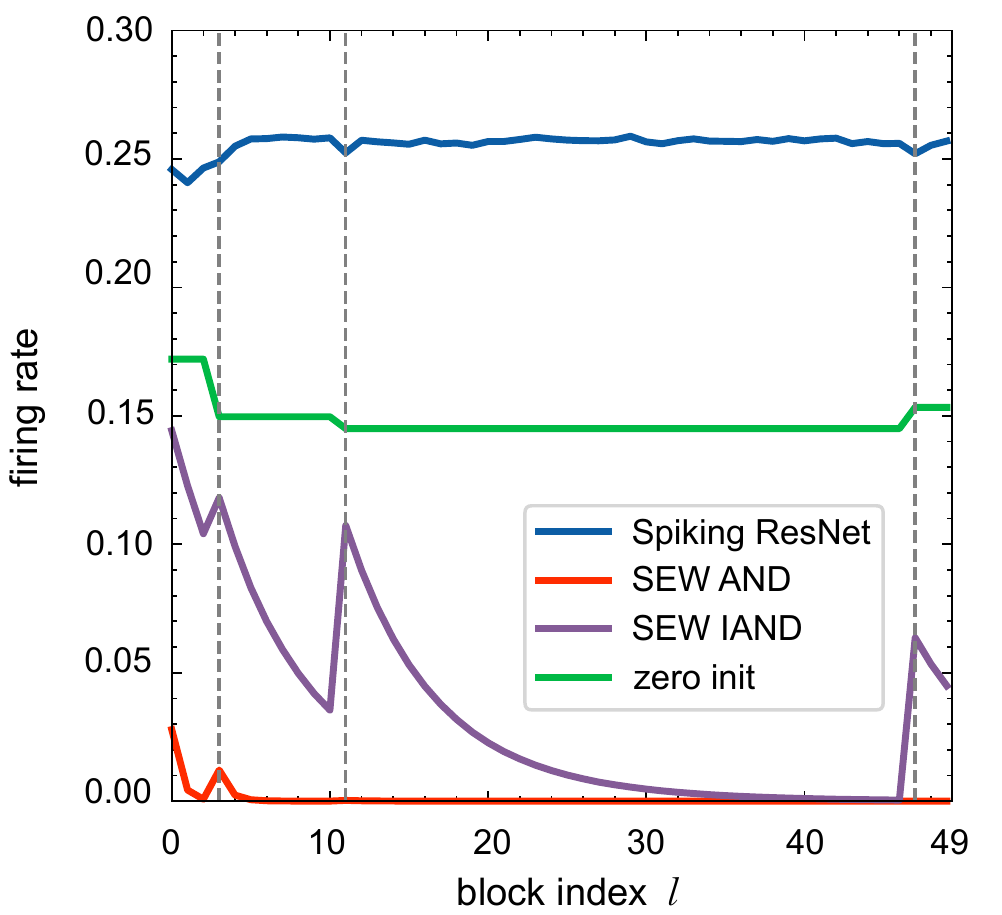}\includegraphics[width=0.3\textwidth,trim=280 0 280 0,clip]{./fig/fig6.pdf}}
		\subfigure[Firing rate of $A^{l}$]{\includegraphics[width=0.3\textwidth,trim=560 0 0 0,clip]{./fig/fig6.pdf}}
		\caption{The initial firing rates of output $O^{l}$ and $A^{l}$ in $l$-th block on 152-layer network.}
		\label{figure:fr gd}
	\end{center}
\end{figure}

As the gradients of SNNs are significantly influenced by firing rates (see Sec.\ref{Gradients in Spiking ResNet with Firing Rates}), we analyze the firing rate firstly.
Fig.~\ref{figure:fr gd}(a) shows the initial firing rate of $l$-th block's output $O^{l}$. The indexes of downsample blocks are marked by vertical dotted lines. The blocks between two adjacent dotted lines represent the identity mapping areas, and have inputs and outputs with the same shape. When using zero initialization, Spiking ResNet, SEW AND ResNet, SEW IAND ResNet, and SEW ADD ResNet have the same firing rates (green curve), which is the \textit{zero init} curve.
Without zero initialization, the silence problem happens in the SEW AND network (red curve), and is relieved by the SEW IAND network (purple curve).  Fig.~\ref{figure:fr gd}(b) shows the firing rate of $A^{l}$, which represents the output of last SN in $l$-th block. It can be found that although the firing rate of $O^{l}$ in SEW ADD ResNet increases linearly in the identity mapping areas, the last SN in each block still maintains a stable firing rate. Note that when $g$ is \textit{ADD}, the output of the SEW block is not binary, and the firing rate is actually the mean value. The SNs of SEW IAND ResNet maintain an adequate firing rate and decay slightly with depth (purple curve), while SNs in deep layers of SEW AND ResNet keep silent (orange curve). The silence problem can be explained as follows. When using \textit{AND}, $O^{l}[t] = {\rm SN}(\mathcal{F}^{l}(O^{l-1}[t])) \land O^{l-1}[t] \leq O^{l-1}[t]$. Since it is hard to keep ${\rm SN}(\mathcal{F}^{l}(O^{l-1}[t])) \equiv 1$ at each time-step $t$, the silence problem may frequently happen in SEW ResNet with \textit{AND} as $g$. Using \textit{IAND} as a substitute of \textit{AND} can relieve this problem because it is easy to keep ${\rm SN}(\mathcal{F}^{l}(O^{l-1}[t])) \equiv 0$ at each time-step $t$.

\begin{figure}[t]
	\begin{center}
		\includegraphics[width=1\textwidth,trim=0 0 0 0,clip]{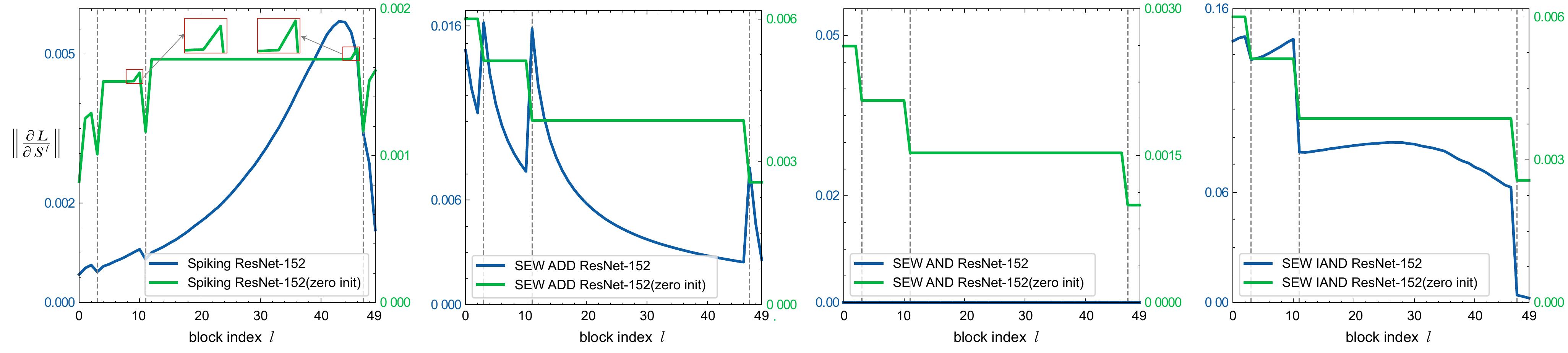}
		\vspace{-0.7cm}
		\caption{Gradient amplitude $\left\| \frac{\partial L}{\partial S^{l}} \right\|$ of $l$-th block when $V_{th}=1, \alpha=2$.} 
		\label{figure:gradient}
		\vspace{-0.5cm}
	\end{center}
\end{figure}
\begin{figure}[t]
	\begin{center}
		\includegraphics[width=1\textwidth,trim=0 0 0 0,clip]{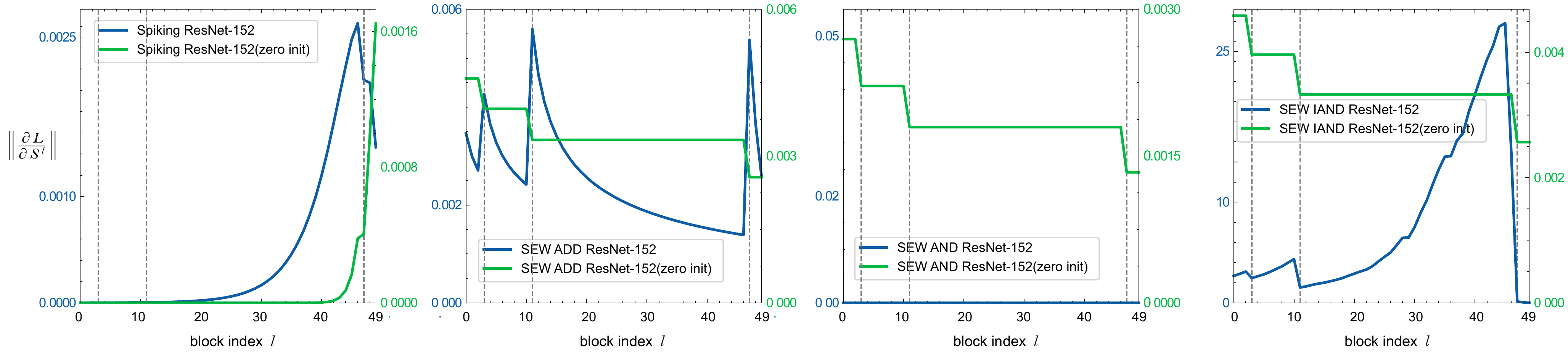}
		\vspace{-0.7cm}
		\caption{Gradient amplitude $\left\| \frac{\partial L}{\partial S^{l}} \right\|$ of $l$-th block when $V_{th}=0.5, \alpha=2$.} 
		\label{figure:gradient van}
		\vspace{-0.5cm}
	\end{center}
\end{figure}
\begin{figure}[t]
	\begin{center}
		\includegraphics[width=1.0\textwidth,trim=0 0 0 0,clip]{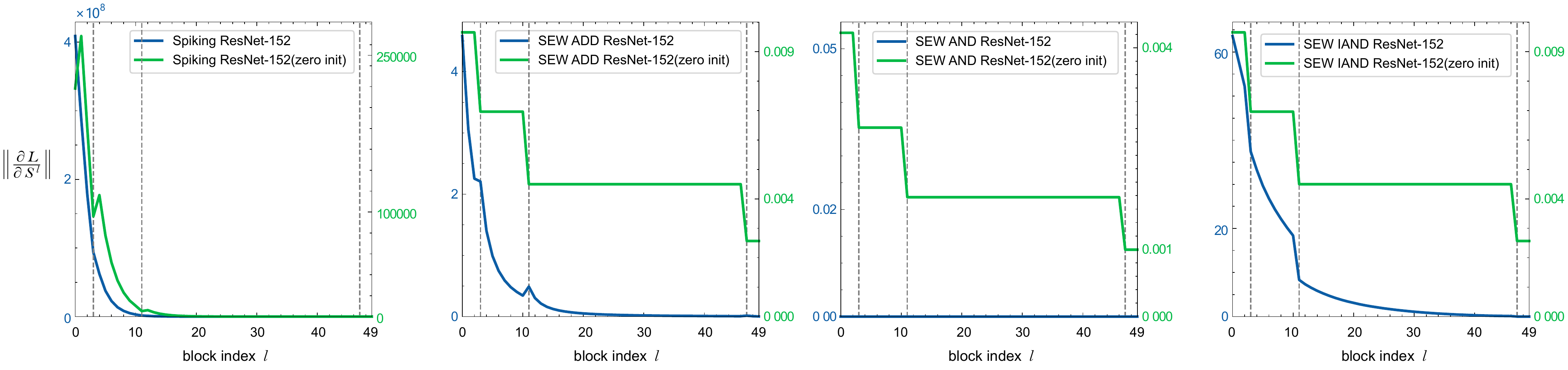}
		\vspace{-0.7cm}
		\caption{Gradient amplitude $\left\| \frac{\partial L}{\partial S^{l}} \right\|$ of $l$-th block when $V_{th}=1, \alpha=3$.} 
		\label{figure:gradient exp}
	\end{center}
\end{figure}

The surrogate gradient function we used in all experiments is $\sigma(x) = \frac{1}{\pi} \arctan(\frac{\pi}{2}\alpha x) + \frac{1}{2}$, thus $\sigma'(x) = \frac{\alpha}{2(1 + (\frac{\pi}{2} \alpha x)^2)}$. When $V_{th}=1, \alpha=2$, the gradient amplitude $\left\| \frac{\partial L}{\partial S^{l}} \right\|$ of each block is shown in Fig.~\ref{figure:gradient}. Note that $\alpha=2$, $\sigma'(x) \leq \sigma'(0) = \sigma'(1 - V_{th}) = 1$ and $\sigma'(0 - V_{th}) = 0.092 < 1$. It can be found that the gradients in Spiking ResNet-152 decay from deeper layers to shallower layers in the identity mapping areas without zero initialization, which is caused by $\sigma'(x) \leq 1$. It is worth noting that the decay also happens in Spiking ResNet-152 with zero initialization. The small convex $\bigwedge$ near the dotted lines is caused by the vanishing gradients of those $S_{j}^{l}[t] = 0$. After these gradients decays to 0 completely, $\left\| \frac{\partial L}{\partial S^{l}} \right\|$ will be a constant because the rest gradients are calculated by $S_{j}^{l}[t] = 1$ and $\sigma'(1 - V_{th}) = 1$, which can also explain why the gradient-index curve is horizontal at some areas. When referring to SEW ResNet-152 with zero initialization, it can be found that all gradient-index curves are similar no matter what $g$ we choose. This is caused by that in the identity mapping areas, $S^{l}$ is constant for all index $l$, and the gradient also becomes a constant as it will not flow through SNs. Without zero initialization, the vanishing gradient happens in the SEW AND ResNet-152, which is caused by the silence problem.  The gradients of SEW ADD, IAND network increase slowly when propagating from deeper layers to shallower layers, due to the adequate firing rates shown in Fig.~\ref{figure:fr gd}.

When $V_{th}=0.5, \alpha=2$, $\sigma'(0 - V_{th}) = \sigma'(1 - V_{th}) = 0.288 < 1$, indicating that transmitting spikes to SNs is prone to causing vanishing gradient, as shown in Fig.~\ref{figure:gradient van}. With zero initialization, the decay in Spiking ResNet-152 is more serious because gradient from $\mathcal{F}^{l}$ can not contribute. The SEW ResNet-152 will not be affected no matter what $g$ we choose. When $V_{th}=1, \alpha=3$, $\sigma'(1 - V_{th}) = 1.5 > 1$, indicating that transmitting spikes to SNs is prone to causing exploding gradient. Fig.~\ref{figure:gradient exp} shows the gradient in this situation. Same with the reason in Fig.~\ref{figure:gradient}, the change of surrogate function will increase gradients of all networks without zero initialization, but not affect SEW ResNet-152 with zero initialization. The Spiking ResNet-152 meets exploding gradient, while this problem in SEW ADD, IAND ResNet-152 is not serious.

\subsection{DVS Gesture Classification} \label{hyp dvsg}
The origin ResNet, which is designed for classifying the complex ImageNet dataset, is too large for the DVS Gesture dataset. Hence, we design a tiny network named 7B-Net, whose structure is \textit{c32k3s1-BN-PLIF-\{SEW Block-MPk2s2\}*7-FC11}. Here c32k3s1 means the convolutional layer with channels 32, kernel size 3, stride 1. MPk2s2 is the max pooling with kernel size 2, stride 2. The symbol \{\}*7 denotes seven repeated structure, and PLIF denotes the Parametric Leaky-Integrate-and-Fire Spiking Neuron with a learnable membrane time constant, which is proposed in \cite{fang2020incorporating} and can be described by Eq.~(\ref{LIFneuron}). See Sec.\ref{Hyper-Parameters} for AER data pre-processing details.

\begin{figure}
	\begin{center}
		\subfigure[Training loss]{\includegraphics[width=0.245\textwidth,trim=0 0 560 0,clip]{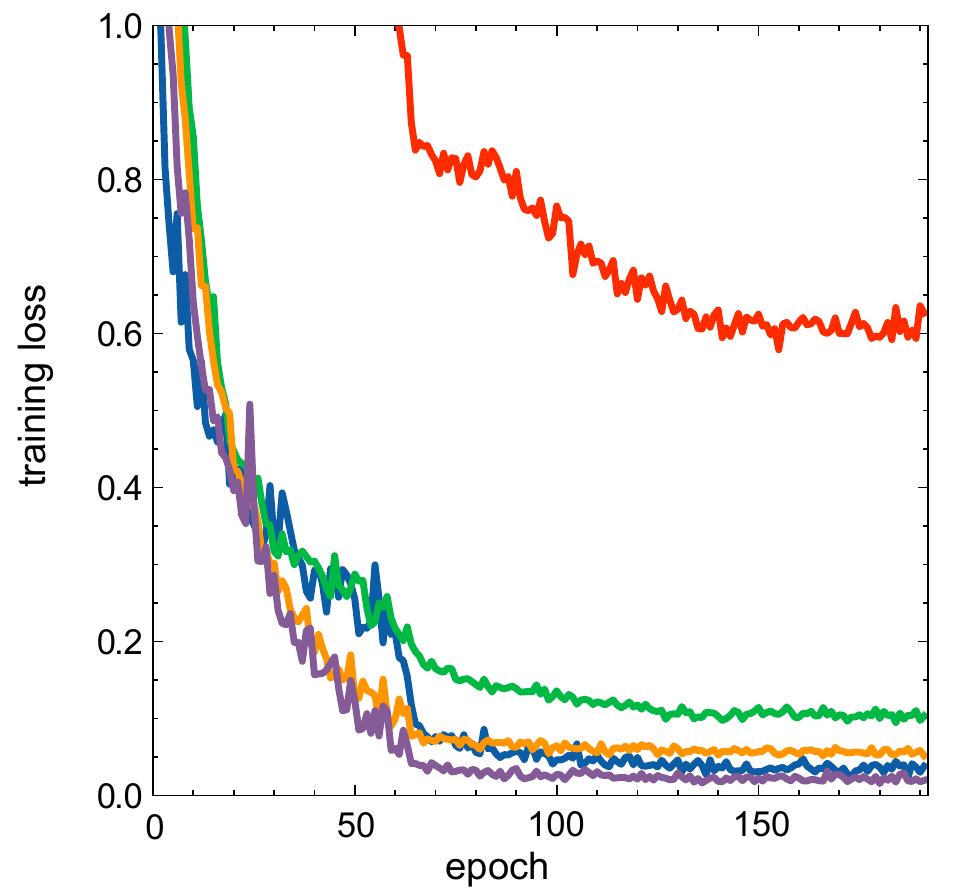}}
		\subfigure[Training accuracy]{\includegraphics[width=0.245\textwidth,trim=280 0 280 0,clip]{./fig/fig5_1.pdf}}
		\subfigure[Test accuracy]{\includegraphics[width=0.245\textwidth,trim=560 0 0 0,clip]{./fig/fig5_1.pdf}}
		\vspace{-0.1cm}
		\caption{Comparison of the training loss, training accuracy and test accuracy on DVS Gesture dataset.}
		\vspace{-0.5cm}
		\label{figure:train curve on DVS Gesture}
	\end{center}
\end{figure}

\textbf{Spiking ResNet \emph{vs.} SEW ResNet.}
We first compare the performance of SEW ResNet with \textit{ADD} element-wise function (SEW ADD ResNet) and Spiking ResNet by replacing SEW blocks with basic blocks. As shown in Fig.~\ref{figure:train curve on DVS Gesture} and Tab.~\ref{tab:acc on DVS Gesture}, although the training loss of Spiking ResNet (blue curve) is lower than SEW ADD ResNet (orange curve), the test accuracy is lower than SEW ADD ResNet (90.97\% v.s. 97.92\%), which implies that Spiking ResNet is easier to overfit than SEW ADD ResNet.

\textbf{Evaluation of different element-wise functions and plain block.}
As the training cost of SNNs on the DVS Gesture dataset is much lower than on ImageNet, we carry out more ablation experiments on the DVS Gesture dataset.
We replace SEW blocks with the plain blocks (no shortcut connection) and test the performance. We also evaluate all kinds of element-wise functions $g$ in Tab.~\ref{tab:g}. Fig.~\ref{figure:train curve on DVS Gesture} shows the training loss and training/test accuracy on DVS Gesture. The sharp fluctuation during early epochs is caused by the large learning rate (see Sec.\ref{Hyper-Parameters}). We can find that the training loss is SEW IAND$<$Spiking ResNet$<$SEW ADD$<$Plain Net$<$SEW AND. Due to the overfitting problem, a lower loss does not guarantee a higher test accuracy.
\begin{wraptable}{r}{8cm}
	\centering
	\scalebox{0.8}
	{
		\begin{tabular}{lcc}
			\hline
			\textbf{Network} & \textbf{Element-Wise Function $g$} & \textbf{Accuracy(\%)}\\
			\hline
			SEW ResNet & ADD & 97.92 \\
			SEW ResNet & IAND & 95.49 \\
			Plain Net & $-$ & 91.67 \\
			Spiking ResNet & $-$ & 90.97 \\
			SEW ResNet & AND & 70.49 \\
			\hline
		\end{tabular}
	}
	\caption{Test accuracy on DVS Gesture. The networks' order is ranked by accuracy.}
	\label{tab:acc on DVS Gesture}
\end{wraptable}
Tab.~\ref{tab:acc on DVS Gesture} shows the test accuracy of all networks. The SEW ADD ResNet gets the highest accuracy than others.

\textbf{Comparisons with State-of-the-art Methods.}
Tab.~\ref{tab:cmp sota on DVS Gesture} compares our network with SOTA methods. It can be found that our SEW ResNet outperforms the SOTA works in accuracy, parameter numbers, and simulating time-steps.
\begin{table}[H]
	\centering
	\scalebox{0.8}
	{
		\begin{tabular}{lccc}
			\hline
			\textbf{Network} & \textbf{Accuracy(\%)} & \textbf{Parameters} & $ \boldsymbol {T}$\\
			\hline
			\textbf{c32k3s1-BN-PLIF-\{SEW Block (c32) -MPk2s2\}*7-FC11 (7B-Net)}& 97.92 & 0.13M & 16\\
			\hline
			\makecell[l]{\{c128k3s1-BN-PLIF-MPk2s2\}*5-DP-\\FC512-PLIF-DP-FC110-PLIF-APk10s10 \cite{fang2020incorporating}} & 97.57 & 1.70M & 20\\
			\hline
			Spiking ResNet-17 with td-BN \cite{zheng2020going} & 96.87 & 11.18M & 40\\
			\hline
			MPk4-c64k3-LIF-c128k3-LIF-APk2-c128k3-LIF-APk2-FC256-LIF-FC11\cite{HE2020108} & 93.40 & 23.23M & 60\\
			\hline
		\end{tabular}
	}
	\vspace{0.1cm}
	\caption{Comparison with the state-of-the-art (SOTA) methods on DVS Gesture dataset.}
	\label{tab:cmp sota on DVS Gesture}
	\vspace{-0.3cm}
\end{table}

\subsection{CIFAR10-DVS Classification}
We also report SEW ResNet on the CIFAR10-DVS dataset, which is obtained by recording the moving images of the CIFAR-10 dataset on a LCD monitor by a DVS camera. As CIFAR10-DVS is more complicated than DVS Gesture, we use the network structure named Wide-7B-Net, which is similar to 7B-Net but with more channels. The structure of Wide-7B-Net is \textit{c64k3s1-BN-PLIF-\{SEW Block (c64)-MPk2s2\}*4-c128k3s1-BN-PLIF-\{SEW Block (c128)-MPk2s2\}*3-FC10}. 
\begin{table}[H]
	\centering
	\scalebox{0.8}
	{
		\begin{tabular}{lccc}
			\hline
			\textbf{Network} & \textbf{Accuracy(\%)} & \textbf{Parameters} & $ \boldsymbol {T}$\\
			\hline
			\bfseries\makecell[l]{c64k3s1-BN-PLIF-\{SEW Block (c64)-MPk2s2\}*4-c128k3s1-\\BN-PLIF-\{SEW Block (c128)-MPk2s2\}*3-FC10 (Wide-7B-Net)}& 64.8, 70.2, 74.4 & 1.19M & 	4, 8, 16\\
			\hline
			\makecell[l]{\{c128k3s1-BN-PLIF-MPk2s2\}*4-DP-FC512-PLIF-DP-\\FC100-PLIF-APk10s10\cite{fang2020incorporating}} & 74.8 & 17.4M & 20\\
			\hline
			Spiking ResNet-19 with td-BN \cite{zheng2020going} & 67.8 & 11.18M & 10\\
			\hline
		\end{tabular}
	}
	\vspace{0.1cm}
	\caption{Comparison with the state-of-the-art (SOTA) methods on CIFAR10-DVS dataset.}
	\label{tab:cmp sota on CIFAR10-DVS}
	\vspace{-0.3cm}
\end{table}

In Tab.\ref{tab:cmp sota on CIFAR10-DVS}, we compare SEW ResNet with the previous Spiking ResNet. One can find that our method achieves better performance (70.2\% v.s. 67.8\%) and fewer time-steps  (8 v.s. 10) than the Spiking ResNet \cite{zheng2020going}. We also compare our method with the state-of-the-art (SOTA) supervised learning methods on CIFAR10-DVS. The accuracy of our Wide-7B-Net is slightly lower than the current SOTA method \cite{fang2020incorporating} (74.4\% v.s. 74.8\%), which uses 1.25 times as many simulation time-steps $T$ (20 v.s. 16) and 14.6 times as many the number of parameters  (17.4M v.s. 1.19M). Moreover, when reducing $T$ shapely to $T=4$, our Wide-7B-Net can still get the accuracy of 64.8\%.

\section{Conclusion}
In this paper, we analyze the previous Spiking ResNet whose residual block mimics the standard block of ResNet, and find that it can hardly implement identity mapping and suffers from the problems of vanishing/exploding gradient. To solve these problems, we propose the SEW residual block and prove that it can implement the residual learning. The experiment results on ImageNet, DVS Gesture, and CIFAR10-DVS datasets show that our SEW residual block solves the degradation problem, and SEW ResNet can achieve higher accuracy by simply increasing the network's depth. Our work may shed light on the learning of ``very deep'' SNNs.

\section{Acknowledgment}
This work is supported by grants from the National Natural Science Foundation of China under contracts No.62027804, No.61825101, and No.62088102.

\bibliographystyle{plain}
\bibliography{ref}

\begin{thebibliography}{10}

\bibitem{amir2017low}
Arnon Amir, Brian Taba, David Berg, Timothy Melano, Jeffrey McKinstry, Carmelo
  Di~Nolfo, Tapan Nayak, Alexander Andreopoulos, Guillaume Garreau, Marcela
  Mendoza, Jeff Kusnitz, Michael Debole, Steve Esser, Tobi Delbruck, Myron
  Flickner, and Dharmendra Modha.
\newblock A low power, fully event-based gesture recognition system.
\newblock In {\em Proceedings of the IEEE Conference on Computer Vision and
  Pattern Recognition (CVPR)}, pages 7243--7252, 2017.

\bibitem{bahdanau2014neural}
Dzmitry Bahdanau, Kyunghyun Cho, and Yoshua Bengio.
\newblock Neural machine translation by jointly learning to align and
  translate.
\newblock In {\em International Conference on Learning Representations (ICLR)},
  2015.

\bibitem{bengio2007scaling}
Yoshua Bengio, Yann LeCun, et~al.
\newblock Scaling learning algorithms towards ai.
\newblock {\em Large-scale Kernel Machines}, 34(5):1--41, 2007.

\bibitem{cao2015spiking}
Yongqiang Cao, Yang Chen, and Deepak Khosla.
\newblock Spiking deep convolutional neural networks for energy-efficient
  object recognition.
\newblock {\em International Journal of Computer Vision}, 113(1):54--66, 2015.

\bibitem{comsa2020temporal}
Iulia~M Comsa, Krzysztof Potempa, Luca Versari, Thomas Fischbacher, Andrea
  Gesmundo, and Jyrki Alakuijala.
\newblock Temporal coding in spiking neural networks with alpha synaptic
  function.
\newblock In {\em International Conference on Acoustics, Speech and Signal
  Processing (ICASSP)}, pages 8529--8533. IEEE, 2020.

\bibitem{deng2021optimal}
Shikuang Deng and Shi Gu.
\newblock Optimal conversion of conventional artificial neural networks to
  spiking neural networks.
\newblock In {\em International Conference on Learning Representations (ICLR)},
  2021.

\bibitem{SpikingJelly}
Wei Fang, Yanqi Chen, Jianhao Ding, Ding Chen, Zhaofei Yu, Huihui Zhou,
  Yonghong Tian, and other contributors.
\newblock Spikingjelly.
\newblock \url{https://github.com/fangwei123456/spikingjelly}.

\bibitem{fang2020incorporating}
Wei Fang, Zhaofei Yu, Yanqi Chen, Timothee Masquelier, Tiejun Huang, and
  Yonghong Tian.
\newblock Incorporating learnable membrane time constant to enhance learning of
  spiking neural networks.
\newblock In {\em Proceedings of the IEEE/CVF International Conference on
  Computer Vision (ICCV)}, pages 2661--2671, 2021.

\bibitem{girshick2014rich}
Ross Girshick, Jeff Donahue, Trevor Darrell, and Jitendra Malik.
\newblock Rich feature hierarchies for accurate object detection and semantic
  segmentation.
\newblock In {\em Proceedings of the IEEE Conference on Computer Vision and
  Pattern Recognition (CVPR)}, pages 580--587, 2014.

\bibitem{goyal2018accurate}
Priya Goyal, Piotr Doll{\'a}r, Ross Girshick, Pieter Noordhuis, Lukasz
  Wesolowski, Aapo Kyrola, Andrew Tulloch, Yangqing Jia, and Kaiming He.
\newblock Accurate, large minibatch sgd: Training imagenet in 1 hour.
\newblock {\em arXiv preprint arXiv:1706.02677}, 2017.

\bibitem{han2020deep}
Bing Han and Kaushik Roy.
\newblock Deep spiking neural network: Energy efficiency through time based
  coding.
\newblock In {\em European Conference on Computer Vision (ECCV)}, pages
  388--404, 2020.

\bibitem{Han_2020_CVPR}
Bing Han, Gopalakrishnan Srinivasan, and Kaushik Roy.
\newblock Rmp-snn: Residual membrane potential neuron for enabling deeper
  high-accuracy and low-latency spiking neural network.
\newblock In {\em Proceedings of the IEEE/CVF Conference on Computer Vision and
  Pattern Recognition (CVPR)}, pages 13558--13567, 2020.

\bibitem{he2015convolutional}
Kaiming He and Jian Sun.
\newblock Convolutional neural networks at constrained time cost.
\newblock In {\em Proceedings of the IEEE/CVF Conference on Computer Vision and
  Pattern Recognition (CVPR)}, pages 5353--5360, 2015.

\bibitem{he2015deep}
Kaiming He, Xiangyu Zhang, Shaoqing Ren, and Jian Sun.
\newblock Deep residual learning for image recognition.
\newblock In {\em Proceedings of the IEEE/CVF Conference on Computer Vision and
  Pattern Recognition (CVPR)}, pages 770--778, 2016.

\bibitem{he2016identity}
Kaiming He, Xiangyu Zhang, Shaoqing Ren, and Jian Sun.
\newblock Identity mappings in deep residual networks.
\newblock In {\em European Conference on Computer Vision (ECCV)}, pages
  630--645. Springer, 2016.

\bibitem{HE2020108}
Weihua He, YuJie Wu, Lei Deng, Guoqi Li, Haoyu Wang, Yang Tian, Wei Ding,
  Wenhui Wang, and Yuan Xie.
\newblock Comparing snns and rnns on neuromorphic vision datasets: Similarities
  and differences.
\newblock {\em Neural Networks}, 132:108--120, 2020.

\bibitem{hu2020spiking}
Yangfan Hu, Huajin Tang, Yueming Wang, and Gang Pan.
\newblock Spiking deep residual network.
\newblock {\em arXiv preprint arXiv:1805.01352}, 2018.

\bibitem{huang2017densely}
Gao Huang, Zhuang Liu, Laurens Van Der~Maaten, and Kilian~Q Weinberger.
\newblock Densely connected convolutional networks.
\newblock In {\em Proceedings of the IEEE/CVF Conference on Computer Vision and
  Pattern Recognition (CVPR)}, pages 4700--4708, 2017.

\bibitem{huh2017gradient}
Dongsung Huh and Terrence~J Sejnowski.
\newblock Gradient descent for spiking neural networks.
\newblock In {\em Advances in Neural Information Processing Systems (NeurIPS)},
  pages 1440--1450, 2018.

\bibitem{hunsberger2015spiking}
Eric Hunsberger and Chris Eliasmith.
\newblock Spiking deep networks with lif neurons.
\newblock {\em arXiv preprint arXiv:1510.08829}, 2015.

\bibitem{10.3389/fnins.2021.629000}
Sungmin Hwang, Jeesoo Chang, Min-Hye Oh, Kyung~Kyu Min, Taejin Jang, Kyungchul
  Park, Junsu Yu, Jong-Ho Lee, and Byung-Gook Park.
\newblock Low-latency spiking neural networks using pre-charged membrane
  potential and delayed evaluation.
\newblock {\em Frontiers in Neuroscience}, 15:135, 2021.

\bibitem{i2016squeezenet}
Forrest~N Iandola, Song Han, Matthew~W Moskewicz, Khalid Ashraf, William~J
  Dally, and Kurt Keutzer.
\newblock Squeezenet: Alexnet-level accuracy with 50x fewer parameters and< 0.5
  mb model size.
\newblock {\em arXiv preprint arXiv:1602.07360}, 2016.

\bibitem{BN}
Sergey Ioffe and Christian Szegedy.
\newblock Batch normalization: Accelerating deep network training by reducing
  internal covariate shift.
\newblock In {\em International Conference on Machine Learning (ICML)}, pages
  448--456, 2015.

\bibitem{kheradpisheh2020temporal}
Saeed~Reza Kheradpisheh and Timoth{\'e}e Masquelier.
\newblock Temporal backpropagation for spiking neural networks with one spike
  per neuron.
\newblock {\em International Journal of Neural Systems}, 30(06):2050027, 2020.

\bibitem{KIM2018373}
Jaehyun Kim, Heesu Kim, Subin Huh, Jinho Lee, and Kiyoung Choi.
\newblock Deep neural networks with weighted spikes.
\newblock {\em Neurocomputing}, 311:373--386, 2018.

\bibitem{kim2020unifying}
Jinseok Kim, Kyungsu Kim, and Jae-Joon Kim.
\newblock Unifying activation- and timing-based learning rules for spiking
  neural networks.
\newblock In {\em Advances in Neural Information Processing Systems (NeurIPS)},
  pages 19534--19544, 2020.

\bibitem{krizhevsky2014weird}
Alex Krizhevsky.
\newblock One weird trick for parallelizing convolutional neural networks.
\newblock {\em arXiv preprint arXiv:1404.5997}, 2014.

\bibitem{krizhevsky2012imagenet}
Alex Krizhevsky, Ilya Sutskever, and Geoffrey~E Hinton.
\newblock Imagenet classification with deep convolutional neural networks.
\newblock In {\em Advances in Neural Information Processing Systems (NeurIPS)},
  pages 1097--1105, 2012.

\bibitem{lecun2015deep}
Yann LeCun, Yoshua Bengio, and Geoffrey Hinton.
\newblock Deep learning.
\newblock {\em Nature}, 521(7553):436--444, 2015.

\bibitem{lee2020enabling}
Chankyu Lee, Syed~Shakib Sarwar, Priyadarshini Panda, Gopalakrishnan
  Srinivasan, and Kaushik Roy.
\newblock Enabling spike-based backpropagation for training deep neural network
  architectures.
\newblock {\em Frontiers in Neuroscience}, 14, 2020.

\bibitem{lee2016training}
Jun~Haeng Lee, Tobi Delbruck, and Michael Pfeiffer.
\newblock Training deep spiking neural networks using backpropagation.
\newblock {\em Frontiers in Neuroscience}, 10:508, 2016.

\bibitem{10.3389/fnins.2017.00309}
Hongmin Li, Hanchao Liu, Xiangyang Ji, Guoqi Li, and Luping Shi.
\newblock Cifar10-dvs: An event-stream dataset for object classification.
\newblock {\em Frontiers in Neuroscience}, 11:309, 2017.

\bibitem{pmlr-v139-li21d}
Yuhang Li, Shikuang Deng, Xin Dong, Ruihao Gong, and Shi Gu.
\newblock A free lunch from ann: Towards efficient, accurate spiking neural
  networks calibration.
\newblock In {\em International Conference on Machine Learning (ICML)}, volume
  139, pages 6316--6325, 2021.

\bibitem{liu2016ssd}
Wei Liu, Dragomir Anguelov, Dumitru Erhan, Christian Szegedy, Scott Reed,
  Cheng-Yang Fu, and Alexander~C Berg.
\newblock Ssd: Single shot multibox detector.
\newblock In {\em European Conference on Computer Vision (ECCV)}, pages 21--37.
  Springer, 2016.

\bibitem{loshchilov2016sgdr}
Ilya Loshchilov and Frank Hutter.
\newblock {SGDR:} stochastic gradient descent with warm restarts.
\newblock In {\em International Conference on Learning Representations (ICLR)},
  2017.

\bibitem{micikevicius2018mixed}
Paulius Micikevicius, Sharan Narang, Jonah Alben, Gregory Diamos, Erich Elsen,
  David Garcia, Boris Ginsburg, Michael Houston, Oleksii Kuchaiev, Ganesh
  Venkatesh, et~al.
\newblock Mixed precision training.
\newblock In {\em International Conference on Learning Representations (ICLR)},
  2018.

\bibitem{mnih2015human}
Volodymyr Mnih, Koray Kavukcuoglu, David Silver, Andrei~A Rusu, Joel Veness,
  Marc~G Bellemare, Alex Graves, Martin Riedmiller, Andreas~K Fidjeland, Georg
  Ostrovski, et~al.
\newblock Human-level control through deep reinforcement learning.
\newblock {\em Nature}, 518(7540):529--533, 2015.

\bibitem{NIPS2014_109d2dd3}
Guido~F Montufar, Razvan Pascanu, Kyunghyun Cho, and Yoshua Bengio.
\newblock On the number of linear regions of deep neural networks.
\newblock In {\em Advances in Neural Information Processing Systems (NeurIPS)},
  pages 2924--2932, 2014.

\bibitem{mostafa2017supervised}
Hesham Mostafa.
\newblock Supervised learning based on temporal coding in spiking neural
  networks.
\newblock {\em IEEE Transactions on Neural Networks and Learning Systems},
  29(7):3227--3235, 2017.

\bibitem{neftci2019surrogate}
Emre~O Neftci, Hesham Mostafa, and Friedemann Zenke.
\newblock Surrogate gradient learning in spiking neural networks: Bringing the
  power of gradient-based optimization to spiking neural networks.
\newblock {\em IEEE Signal Processing Magazine}, 36(6):51--63, 2019.

\bibitem{PYTORCH}
Adam Paszke, Sam Gross, Francisco Massa, Adam Lerer, James Bradbury, Gregory
  Chanan, Trevor Killeen, Zeming Lin, Natalia Gimelshein, Luca Antiga, Alban
  Desmaison, Andreas Kopf, Edward Yang, Zachary DeVito, Martin Raison, Alykhan
  Tejani, Sasank Chilamkurthy, Benoit Steiner, Lu~Fang, Junjie Bai, and Soumith
  Chintala.
\newblock Pytorch: An imperative style, high-performance deep learning library.
\newblock In {\em Advances in Neural Information Processing Systems (NeurIPS)},
  pages 8026--8037, 2019.

\bibitem{rathi2020dietsnn}
Nitin Rathi and Kaushik Roy.
\newblock Diet-snn: Direct input encoding with leakage and threshold
  optimization in deep spiking neural networks.
\newblock {\em arXiv preprint arXiv:2008.03658}, 2020.

\bibitem{rathi2020enabling}
Nitin Rathi, Gopalakrishnan Srinivasan, Priyadarshini Panda, and Kaushik Roy.
\newblock Enabling deep spiking neural networks with hybrid conversion and
  spike timing dependent backpropagation.
\newblock In {\em International Conference on Learning Representations (ICLR)},
  2020.

\bibitem{redmon2016you}
Joseph Redmon, Santosh Divvala, Ross Girshick, and Ali Farhadi.
\newblock You only look once: Unified, real-time object detection.
\newblock In {\em Proceedings of the IEEE Conference on Computer Vision and
  Pattern Recognition (CVPR)}, pages 779--788, 2016.

\bibitem{roy2019towards}
Kaushik Roy, Akhilesh Jaiswal, and Priyadarshini Panda.
\newblock Towards spike-based machine intelligence with neuromorphic computing.
\newblock {\em Nature}, 575(7784):607--617, 2019.

\bibitem{Bodo2017Conversion}
Bodo Rueckauer, Iulia-Alexandra Lungu, Yuhuang Hu, Michael Pfeiffer, and
  Shih-Chii Liu.
\newblock Conversion of continuous-valued deep networks to efficient
  event-driven networks for image classification.
\newblock {\em Frontiers in Neuroscience}, 11:682, 2017.

\bibitem{russakovsky2015imagenet}
Olga Russakovsky, Jia Deng, Hao Su, Jonathan Krause, Sanjeev Satheesh, Sean Ma,
  Zhiheng Huang, Andrej Karpathy, Aditya Khosla, Michael Bernstein, et~al.
\newblock Imagenet large scale visual recognition challenge.
\newblock {\em International Journal of Computer Vision}, 115(3):211--252,
  2015.

\bibitem{samadzadeh2021convolutional}
Ali Samadzadeh, Fatemeh Sadat~Tabatabaei Far, Ali Javadi, Ahmad Nickabadi, and
  Morteza~Haghir Chehreghani.
\newblock Convolutional spiking neural networks for spatio-temporal feature
  extraction.
\newblock {\em arXiv preprint arXiv:2003.12346}, 2020.

\bibitem{sengupta2019going}
Abhronil Sengupta, Yuting Ye, Robert Wang, Chiao Liu, and Kaushik Roy.
\newblock Going deeper in spiking neural networks: Vgg and residual
  architectures.
\newblock {\em Frontiers in Neuroscience}, 13:95, 2019.

\bibitem{shrestha2018slayer}
Sumit~Bam Shrestha and Garrick Orchard.
\newblock Slayer: Spike layer error reassignment in time.
\newblock In {\em Advances in Neural Information Processing Systems (NeurIPS)},
  pages 1419--1428, 2018.

\bibitem{silver2016mastering}
David Silver, Aja Huang, Chris~J Maddison, Arthur Guez, Laurent Sifre, George
  Van Den~Driessche, Julian Schrittwieser, Ioannis Antonoglou, Veda
  Panneershelvam, Marc Lanctot, et~al.
\newblock Mastering the game of go with deep neural networks and tree search.
\newblock {\em Nature}, 529(7587):484--489, 2016.

\bibitem{simonyan2015deep}
Karen Simonyan and Andrew Zisserman.
\newblock Very deep convolutional networks for large-scale image recognition.
\newblock In {\em International Conference on Learning Representations (ICLR)},
  2015.

\bibitem{srivastava2015highway}
Rupesh~Kumar Srivastava, Klaus Greff, and J{\"u}rgen Schmidhuber.
\newblock Highway networks.
\newblock {\em arXiv preprint arXiv:1505.00387}, 2015.

\bibitem{stockl2021optimized}
Christoph St{\"o}ckl and Wolfgang Maass.
\newblock Optimized spiking neurons can classify images with high accuracy
  through temporal coding with two spikes.
\newblock {\em Nature Machine Intelligence}, 3(3):230--238, 2021.

\bibitem{szegedy2015going}
Christian Szegedy, Wei Liu, Yangqing Jia, Pierre Sermanet, Scott Reed, Dragomir
  Anguelov, Dumitru Erhan, Vincent Vanhoucke, and Andrew Rabinovich.
\newblock Going deeper with convolutions.
\newblock In {\em Proceedings of the IEEE/CVF Conference on Computer Vision and
  Pattern Recognition (CVPR)}, pages 1--9, 2015.

\bibitem{TAVANAEI201947}
Amirhossein Tavanaei, Masoud Ghodrati, Saeed~Reza Kheradpisheh, Timoth{\'e}e
  Masquelier, and Anthony Maida.
\newblock Deep learning in spiking neural networks.
\newblock {\em Neural Networks}, 111:47--63, 2019.

\bibitem{vaswani2017attention}
Ashish Vaswani, Noam Shazeer, Niki Parmar, Jakob Uszkoreit, Llion Jones,
  Aidan~N Gomez, \L~ukasz Kaiser, and Illia Polosukhin.
\newblock Attention is all you need.
\newblock In {\em Advances in Neural Information Processing Systems (NeurIPS)},
  pages 5998--6008, 2017.

\bibitem{wu2018STBP}
Yujie Wu, Lei Deng, Guoqi Li, Jun Zhu, and Luping Shi.
\newblock Spatio-temporal backpropagation for training high-performance spiking
  neural networks.
\newblock {\em Frontiers in Neuroscience}, 12:331, 2018.

\bibitem{xie2016aggregated}
Saining Xie, Ross Girshick, Piotr Dollár, Zhuowen Tu, and Kaiming He.
\newblock Aggregated residual transformations for deep neural networks.
\newblock In {\em Proceedings of the IEEE/CVF Conference on Computer Vision and
  Pattern Recognition (CVPR)}, pages 5987--5995, 2017.

\bibitem{10.1007/978-3-030-36718-3_15}
Fu~Xing, Ye~Yuan, Hong Huo, and Tao Fang.
\newblock Homeostasis-based cnn-to-snn conversion of inception and residual
  architectures.
\newblock In {\em International Conference on Neural Information Processing},
  pages 173--184. Springer, 2019.

\bibitem{yin2020effective}
Bojian Yin, Federico Corradi, and Sander~M Boht{\'e}.
\newblock Effective and efficient computation with multiple-timescale spiking
  recurrent neural networks.
\newblock In {\em International Conference on Neuromorphic Systems}, pages
  1--8, 2020.

\bibitem{Zenke2020.06.29.176925}
Friedemann Zenke and Tim~P Vogels.
\newblock The remarkable robustness of surrogate gradient learning for
  instilling complex function in spiking neural networks.
\newblock {\em Neural Computation}, 33(4):899--925, 2021.

\bibitem{zhang2020temporal}
Wenrui Zhang and Peng Li.
\newblock Temporal spike sequence learning via backpropagation for deep spiking
  neural networks.
\newblock In {\em Advances in Neural Information Processing Systems (NeurIPS)},
  pages 12022--12033, 2020.

\bibitem{zheng2020going}
Hanle Zheng, Yujie Wu, Lei Deng, Yifan Hu, and Guoqi Li.
\newblock Going deeper with directly-trained larger spiking neural networks.
\newblock In {\em Proceedings of the AAAI Conference on Artificial
  Intelligence}, volume~35, pages 11062--11070, 2021.

\bibitem{zhou2019temporal}
Shibo Zhou, Xiaohua Li, Ying Chen, Sanjeev~T. Chandrasekaran, and Arindam
  Sanyal.
\newblock Temporal-coded deep spiking neural network with easy training and
  robust performance.
\newblock In {\em Proceedings of the AAAI Conference on Artificial
  Intelligence}, volume~35, pages 11143--11151, 2021.

\bibitem{zimmer2019technical}
Romain Zimmer, Thomas Pellegrini, Srisht~Fateh Singh, and Timoth{\'e}e
  Masquelier.
\newblock Technical report: supervised training of convolutional spiking neural
  networks with pytorch.
\newblock {\em arXiv preprint arXiv:1911.10124}, 2019.

\end{thebibliography}

\appendix

\section{Appendix}

\subsection{Hyper-Parameters} \label{Hyper-Parameters}
For all datasets, the surrogate gradient function is $\sigma(x) = \frac{1}{\pi} \arctan(\frac{\pi}{2}\alpha x) + \frac{1}{2}$, thus $\sigma'(x) = \frac{\alpha}{2(1 + (\frac{\pi}{2} \alpha x)^2)}$, where $\alpha$ is the slope parameter. We set $\alpha=2$, $V_{reset} = 0$ and $V_{th} = 1$ for all neurons. The optimizer is SGD with momentum 0.9. As recommended by \cite{Zenke2020.06.29.176925}, we detach $S[t]$ in the neuronal reset Eq.~(\ref{neural reset}) in the backward computational graph to improve performance. We use the mixed precision training \cite{micikevicius2018mixed}, which will accelerate training and decrease memory consumption, but may cause slightly lower accuracy than using full precision training. The hyper-parameters of the SNNs for different datasets are shown in Tab.~\ref{tab:hyp for imagenet and dvsg}. Tab.~\ref{tab:lr for dvsg} shows the learning rates of the SNNs with different element-wise functions for DVS Gesture. The data pre-processing methods for three datasets are as following:
\paragraph{ImageNet}
The data augmentation methods used in \cite{he2015deep} are also applied in our experiments. A 224×224 crop is randomly sampled from an image or its horizontal flip with data normalization for train samples. A 224×224 resize and central crop with data normalization is applied for test samples. 
\paragraph{DVS128 Gesture}
We use the same AER data pre-processing method as \cite{fang2020incorporating}, and utilize \textit{random temporal delete} to relieve overfitting, which is illustrated in Sec.~\ref{Random Temporal Delete}.

\paragraph{CIFAR10-DVS}
We use the same AER data pre-processing method as DVS128 Gesture. We do not use \textit{random temporal delete} because CIFAR10-DVS is obtained by static images.

\begin{table}[h]
	\centering
	\scalebox{0.9}
	{
		\begin{tabular}{llllllll}
			\hline
			\textbf{Dataset}& \textbf{Learning Rate Scheduler} &\textbf{Epoch} &$ \boldsymbol {lr}$ & \textbf{Batch Size} & $\boldsymbol {T}$ &$\boldsymbol {n_{gpu}}$\\
			\hline
			ImageNet& Cosine Annealing \cite{loshchilov2016sgdr}, $T_{max}=320$ & 320 &0.1 & 32 & 4 & 8\\
			\hline
			DVS Gesture& Step, $T_{step}=64. \gamma=0.1$ & 192 &0.1 & 16 & 16 & 1\\
			\hline
			CIFAR10-DVS& Cosine Annealing, $T_{max}=64$ & 64 &0.01 & 16 & 4, 8, 16 & 1\\
			\hline
		\end{tabular}
	}
	\caption{Hyper-parameters of the SNNs for three datasets.}
	\label{tab:hyp for imagenet and dvsg}
\end{table}

\begin{figure}
	\begin{center}
		\includegraphics[width=1.0\textwidth,trim=0 0 0 0,clip]{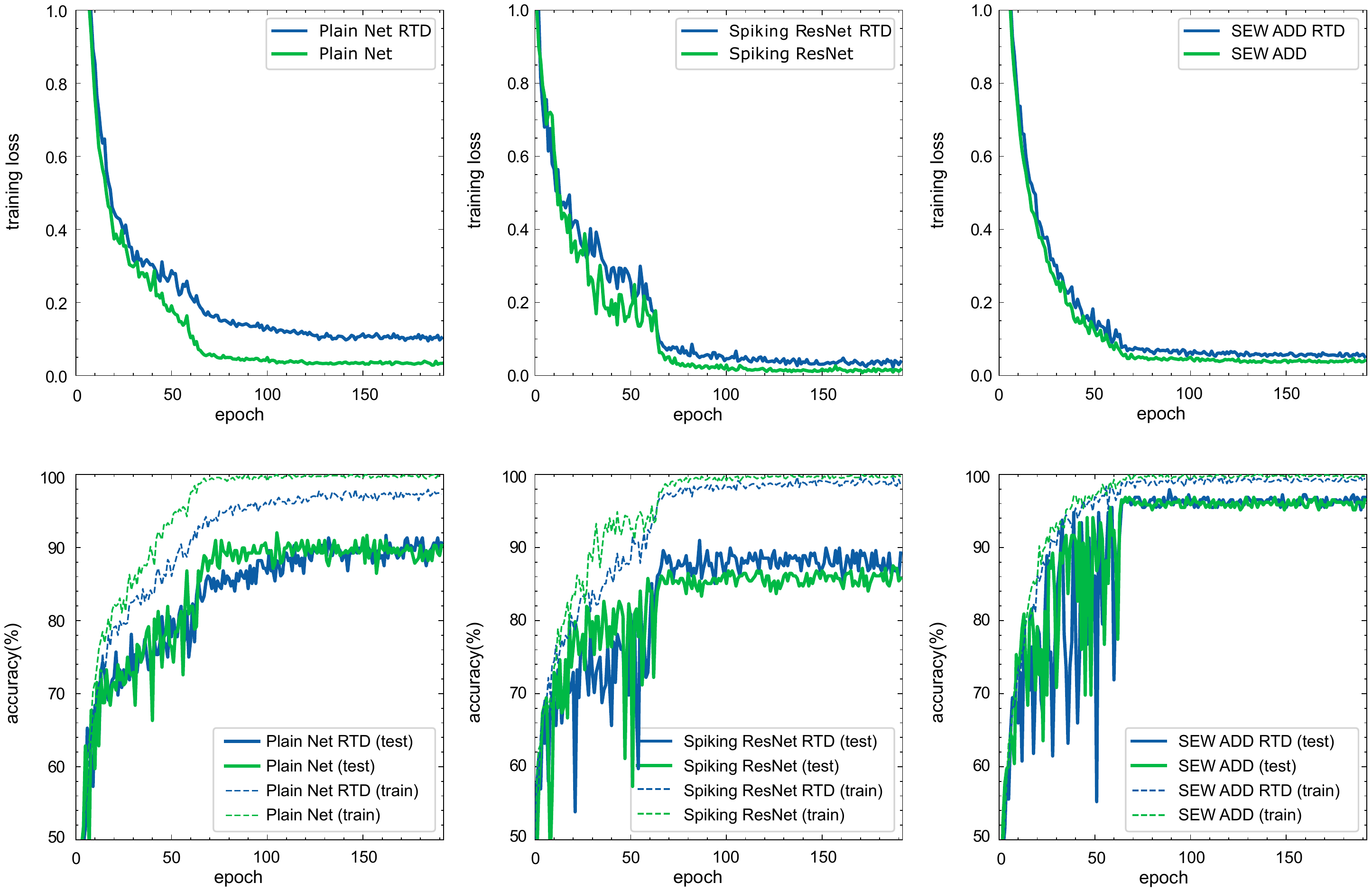}
		\caption{Comparison of training loss and training/test accuracy with/without random temporal delete (RTD).} 
		\label{figure:cmp aug}
	\end{center}
\end{figure}

\begin{table}
	\centering
	\scalebox{1.0}
	{
		\begin{tabular}{ccc}
			\hline
			\textbf{Network} & \textbf{Element-Wise Function $g$} & \textbf{Learning Rate}\\
			\hline
			SEW ResNet & ADD & 0.001 \\
			SEW ResNet & AND & 0.03 \\
			SEW ResNet & IAND & 0.063 \\
			Spiking ResNet & - & 0.1 \\
			Plain Net & - & 0.005 \\
			\hline
		\end{tabular}
	}
	\caption{Learning rates of the SNNs for DVS Gesture.}
	\label{tab:lr for dvsg}
\end{table}

\subsection{Random Temporal Delete} \label{Random Temporal Delete}
To reduce overfitting, we propose a simple data augmentation method called \textit{random temporal delete} for sequential data. Denote the sequence length as $T$, we randomly delete $T - T_{train}$ slices in the origin sequence and use $T_{train}$ slices during training. During inference we use the whole sequence, that is, $T_{test} = T$. We set $T_{train}=12, T=16$ in all experiments on DVS Gesture.

Fig.~\ref{figure:cmp aug} compares the training loss and training/test accuracy of Plain Net, Spiking ResNet, and SEW ResNet with or without \textit{random temporal delete} (RTD). Here the element-wise function $g$ is \textit{ADD}. It can be found that the network with RTD  has higher training loss and lower training accuracy than the network without RTD,
because RTD can increase the difficulty of training. The test accuracy of the network with RTD is higher than that without RTD, showing that RTD will reduce overfitting. The results on the three networks are consistent, indicating that RTD is a general sequential data augmentation method.

\subsection{Firing rates on DVS Gesture}
\begin{figure}
	\begin{center}
		\subfigure[Firing rates of $A^{l}$ in each block on DVS Gesture Gesture]{\includegraphics[width=0.8\textwidth,trim=0 220 0 0,clip]{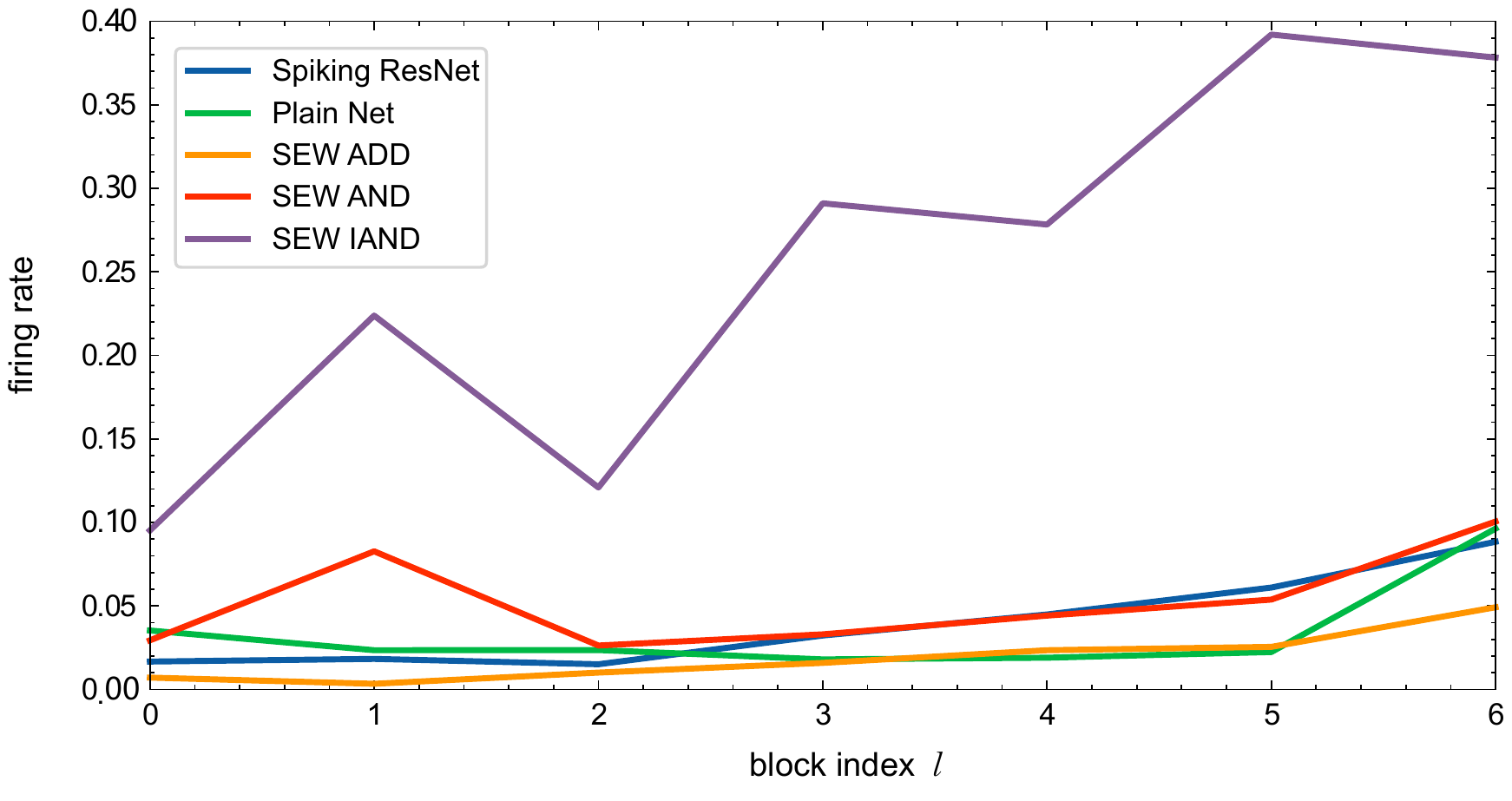}}
		\subfigure[Firing rates of the output $O^{l}$ in each block on DVS Gesture Gesture]{\includegraphics[width=0.8\textwidth,trim=0 0 0 260,clip]{./fig/fig10.pdf}}
		\caption{ Firing rates of the last SN and the output $O^{l}$ in each block of 7B-Net on DVS
			Gesture.} 
		\label{figure:fr on dvsg}
	\end{center}
\end{figure}
Fig.~\ref{figure:fr on dvsg}(a) shows the firing rates of $A^{l}$ in each block from 7B-Net for DVS Gesture. Note that if $g$ is \textit{AND}, the SEW block gets closer to identity mapping when the firing rate approaches 1, while for other $g$, the SEW block becomes identity mapping when the firing rate approaches 0. When all SEW blocks become identity mapping, the 7B-Net will become \textit{c32k3s1-BN-PLIF-\{MPk2s2\}*7-FC11}, which is a too simple network to cause underfitting. Thus, the SEW blocks in 7B-Net are not necessary to be identity mapping. Fig.~\ref{figure:fr on dvsg}(b) shows the firing rates of each block's output $O^{l}$. The firing rates do not strictly decrease with block index increases as blocks are connected by max pooling, which squeezes sparse spikes and increases the firing rate. It can be found that the blocks in SEW AND network have the lowest firing rates. The blocks in SEW IAND network have higher firing rates than those of SEW AND network, and the SEW IAND network has much higher accuracy than the SEW AND network (95.49\% v.s. 70.49\%), indicating that using \textit{IAND} to replace \textit{AND} can relieve the silence problem discussed in Sec.\ref{Gradients Check on ResNet-152 Structure.}.

\subsection{Gradients in Spiking ResNet with Firing Rates} \label{Gradients in Spiking ResNet with Firing Rates}
The gradients of SNNs are affected by firing rates, which is the reason why we analyze the firing rates before gradients in Sec.\ref{Gradients Check on ResNet-152 Structure.}. Consider a spiking ResNet with $k$ sequential blocks to transmit $S^{l}[t]$, and the identity mapping condition is met, e.g., the spiking neurons are the IF neurons with $0 < V_{th} \leq 1$, then we have $S^{l}[t] = S^{l+1}[t] = ... = S^{l+k-1}[t] = O^{l+k-1}[t]$. We get
\begin{align}
	\frac{{\partial O_{j}^{l}[t]}}{{\partial S_{j}^{l}[t]}} &= \frac{{\partial {\rm SN}(S_{j}^{l}[t])}}{{\partial S_{j}^{l}[t]}} = \Theta'(S_{j}^{l}[t] - V_{th})\\
	\frac{\partial L}{\partial S_{j}^{l}[t]} &= \frac{\partial L}{\partial O_{j}^{l}[t]}\Theta'(S_{j}^{l}[t] - V_{th}).
\end{align}
Then the gradients between two adjacent blocks are
\begin{align}
	\frac{\partial L}{\partial O^{l+i}} &= \frac{\partial L}{\partial O^{l+i+1}}\Theta'(S^{l+i+1} - V_{th}).
\end{align}
Denote the number of neurons as $N$, the firing rate of $S^{l}$ as $\Phi = \frac{\sum_{j=0}^{N-1}\sum_{t=0}^{T-1}S_{j}^{l}[t]}{NT}$, then
\begin{align}
	\left\|\frac{\partial L}{\partial S^{l}}\right\| = \left\|\frac{\partial L}{\partial O^{l+k-1}}\right\| \cdot \left\|\prod_{i=0}^{k-1}\Theta'(S^{l+i} - V_{th})\right\|,
\end{align}
where
\begin{align*}
	\left\|\prod_{i=0}^{k-1}\Theta'(S^{l+i} - V_{th})\right\| = \sqrt{NT\Phi(\Theta'(1 - V_{th}))^{2k} + NT(1-\Phi)(\Theta'(0 - V_{th}))^{2k}}\\ \rightarrow
	\begin{cases}
		\sqrt{NT}, &\Theta'(1 - V_{th})=1, \Theta'(0 - V_{th})=1\\
		\sqrt{NT\Phi}, &\Theta'(1 - V_{th})=1, \Theta'(0 - V_{th})<1\\
		\sqrt{NT(1 - \Phi)}, &\Theta'(1 - V_{th})<1, \Theta'(0 - V_{th})=1\\
		0, &\Theta'(1 - V_{th})<1, \Theta'(0 - V_{th})<1\\
		+\infty, &\Theta'(1 - V_{th})>1~or~\Theta'(0 - V_{th})>1.
	\end{cases}
\end{align*}
\subsection{0/1 Gradients Experiments}
As the analysis in Sec.\ref{Spiking ResNet suffers from the problems of vanishing/exploding gradient} shows, the vanishing/exploding gradient problems are easy to happen in Spiking ResNet because of accumulative multiplication. A potential solution is to set $\Theta'(0 - V_{th}) = \Theta'(1 - V_{th})=1$. Specifically, we have trained the Spiking ResNet on ImageNet by setting $V_{th}=0.5$ and $\sigma'(x) = \frac{1 + \frac{\pi^2}{4}}{1 + (\pi x)^{2}}$ in the last SN of each block to make sure that $\Theta'(0 - V_{th}) = \Theta'(1 - V_{th})=1$. However, this network will not converge, which may be caused by that SNNs are sensitive to surrogate functions.

\cite{zheng2020going} uses the Rectangular surrogate function $\sigma'(x)=\frac{1}{a}{\rm sign}(|x|<\frac{a}{2})$. If we set $a=1$, then $\sigma'(x) \in \{0,1\}$. According to Eq.(\ref{block snn gradient}), using this surrogate function can avoid the gradient exploding/vanishing problems in Spiking ResNet. We compare different surrogate functions, including Rectangular ($\sigma'(x)={\rm sign}(|x|<\frac{1}{2})$), ArcTan ($\sigma'(x)=\frac{1}{1 + (\pi x)^2})$ and Constant 1 ($\sigma' (x) \equiv 1$), in the SNNs on CIFAR-10. Note that we aim to evaluate 0/1 gradients, rather than achieve SOTA accuracy. Hence, we use a lightweight network, whose structure is \textit{c32k3s1-BN-IF-\{\{SEW Block (c32)\}*2-MPk2s2\}*5-FC10}. We use \textit{ADD} as $g$ in SEW blocks. We also compare with Spiking ResNet by replacing SEW blocks with basic blocks. The results are shown in Tab.\ref{tab:cmp sg in cf10}. The learning rates for each surrogate function are fine-tuned.

\begin{table}
	\centering
	\scalebox{1.0}
	{
		\begin{tabular}{ccc}
			\hline
			\textbf{Surrogate function} & \textbf{SEW ResNet (ADD)} & \textbf{Spiking ResNet}\\
			\hline
			ArcTan & 0.8263 & 0.7733 \\
			Rectangular & 0.8256 & 0.6601 \\
			Constant 1 & 0.1256 & 0.1 \\
			\hline
		\end{tabular}
	}
	\caption{Test accuracy of SEW ADD ResNet and Spiking ResNet on CIFAR-10 with different surrogate functions.}
	\label{tab:cmp sg in cf10}
\end{table}

Tab.\ref{tab:cmp sg in cf10} indicates that the choice of surrogate function has a considerable influence on the SNN's performance. Although Rectangular and Constant 1 can avoid the gradient exploding/vanishing problems in Eq.(\ref{block snn gradient}), they still cause lower accuracy or even make the optimization not converges. Tab.\ref{tab:cmp sg in cf10} also shows that the SEW ResNet is more robust to the surrogate gradient as it always has higher accuracy than the Spiking ResNet with the same surrogate function.

\subsection{Reproducibility}
All experiments are implemented with SpikingJelly~\cite{SpikingJelly}, which is an open-source deep learning framework for SNNs based on PyTorch~\cite{PYTORCH}. Source codes are available at \url{https://github.com/fangwei123456/Spike-Element-Wise-ResNet}. To maximize reproducibility, we use identical seeds in all codes.

\end{document}